\documentclass[lettersize,journal]{IEEEtran}
\usepackage{amsmath,amsfonts}
\usepackage{algorithmic}
\usepackage{algorithm}
\usepackage{array}
\usepackage[caption=false,font=normalsize,labelfont=sf,textfont=sf]{subfig}
\usepackage{textcomp}
\usepackage{stfloats}
\usepackage{url}
\usepackage{verbatim}
\usepackage{graphicx}
\usepackage{cite}
\usepackage{dsfont}
\usepackage{amstext}
\usepackage{amsmath}
\usepackage{amssymb}
\usepackage{multirow}
\usepackage{color,xcolor,colortbl}
\usepackage{soul}

\definecolor{myblue}{RGB}{186,215,233}
\definecolor{myred}{RGB}{250,162,159}

\usepackage{amsthm}
\theoremstyle{plain}
\newtheorem{theorem}{Theorem}[section]
\newtheorem{proposition}[theorem]{Proposition}

\theoremstyle{definition}

\theoremstyle{remark}
\newtheorem{remark}[theorem]{Remark}
\usepackage[textsize=tiny]{todonotes}

\newcommand{\tabincell}[2]{\begin{tabular}{@{}#1@{}}#2\end{tabular}}  

\definecolor{shadecolor}{rgb}{0.92,0.92,0.92}
\usepackage{framed}

\begin{document}
	
	\title{Rebalanced Zero-shot Learning}
	
	\author{
		Zihan Ye,~\IEEEmembership{Student Member,~IEEE,}
		\thanks{$\dagger$: Corresponding author.\\
			Zihan Ye and Xiaobo Jin are with School of Advanced Technology, Xi'an Jiaotong Liverpool University.}
		Guanyu Yang,
		\thanks{Guanyu Yang and Kaizhu Huang are with Data Science Research Center and Division of Natural and Applied Sciences, Duke Kunshan University.}
		Xiaobo Jin$^\dagger$,
		Youfa Liu,
		\thanks{Youfa Liu is with College of Informatics, Huazhong Agricultural University.}
		Kaizhu Huang,~\IEEEmembership{Senior Member,~IEEE}
	}
	
	
	\maketitle
	
	\begin{abstract}
		Zero-shot learning (ZSL) aims to identify unseen classes with zero samples during training.
		Broadly speaking, present ZSL methods usually adopt class-level semantic labels and compare them with instance-level semantic predictions to infer unseen classes.
		However, we find that such existing models mostly produce imbalanced semantic predictions, i.e. these models could perform precisely for some semantics, but  may not for others. To address the drawback, we aim to introduce an imbalanced learning framework into ZSL. However, we find that imbalanced ZSL has two unique challenges: (1) Its imbalanced predictions are highly correlated with the value of semantic labels rather than the number of samples as typically considered in the traditional imbalanced learning; (2) Different semantics follow quite different error distributions between classes. To mitigate these issues, we first formalize ZSL as an imbalanced regression problem  which offers empirical evidences to interpret how semantic labels lead to imbalanced semantic predictions. We then propose a re-weighted loss termed Re-balanced Mean-Squared Error (ReMSE), which tracks the mean and variance of error distributions, thus ensuring rebalanced learning across classes. As a major contribution, we conduct a series of analyses showing that ReMSE is theoretically well established. Extensive experiments demonstrate that the proposed method effectively alleviates the imbalance in semantic prediction and outperforms many state-of-the-art ZSL methods. Our code is available at \textcolor{magenta}{https://github.com/FouriYe/ReZSL-TIP23}.
	\end{abstract}

	\begin{IEEEkeywords}
		Zero-shot learning, Imbalanced regression, Semantic prediction, Reweighting strategy
	\end{IEEEkeywords}

	\section{Introduction}
	
	Generalization from limited data is the basic cognitive ability of intelligence~\cite{sims2018efficient}. To achieve class-level generalization abilities, researchers develop zero-shot learning (ZSL)~\cite{xian2018zero}, which aims to identify unseen classes without any available images during training. ZSL can also be extended to a more general setting called  generalized ZSL (GZSL) which tries to identify both seen and unseen classes at test time~\cite{chao2016empirical}.
	
	\begin{figure}[htbp]
		\centering
		\includegraphics[width=0.95\linewidth]{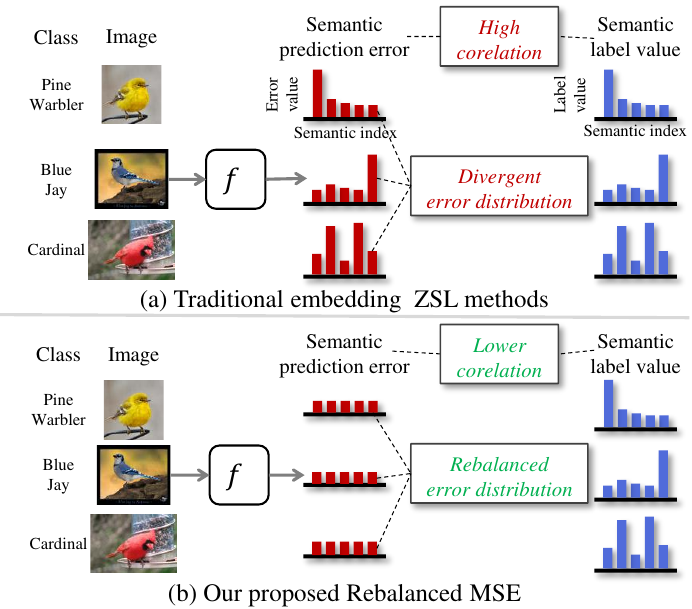}
		\caption{Comparison between traditional embedding ZSL methods and our Re-balanced MSE (ReMSE). Traditional embedding ZSL methods generally produce imbalanced error losses on semantic prediction, which have a high co-relation with label value. Our ReMSE could decease the co-relation and obtain a re-balanced error distribution.}
		\label{fig:banner}
	\end{figure}
	
	Generally, existing ZSL methods utilize various semantic labels, e.g. word2vec~\cite{xian2018zero}, attributes~\cite{lampert2013attribute}, and even auto-mining visually-grounded embeddings~\cite{xu2022vgse} as auxiliary information, to obtain knowledge that transfers from seen to unseen classes.
	Benefiting from these semantic labels, researchers can build a learnable mapping between visual space and semantic space and classify unseen classes in the mapping space, which is called embedding ZSL.
	Recently, ZSL methods have mainly focused on exploiting various complex modules and architectures, such as GCN~\cite{xie2020region} and Transformer~\cite{chen2022transzero} to extract powerful visual features and improve visual-semantic interaction. However, high computational demands caused by such a complex structure pose challenges in practical applications.

	In this work, we address the zero-shot problem from a brand new perspective, namely how semantic labels would affect the performance of ZSL methods.
	Specifically, we find that almost all existing ZSL models suffer from imbalanced semantic prediction, i.e. the model can accurately predict some semantics but may not for others.
	We argue that such problem could significantly limit the performance of ZSL.
	In fact, treating every semantic fairly is a prominent topic in current visual recognition systems~\cite{ramaswamy2021fair, park2022fair}.
	Essentially, an ideal ZSL model should treat all semantic labels with equal importance.
	
	
	However, borrowing the imbalanced regression defined in~\cite{yang2021delving},  we find that such imbalanced semantic prediction in ZSL unfortunately suffers from two unique challenges. On one hand, as shown in Fig.~\ref{fig:banner} (a),  the imbalanced predictions in ZSL are highly correlated with the value of semantic labels, rather than the number of samples as seen in traditional imbalanced learning~\cite{yang2021delving,ren2021balanced,hu2020learning}; on the other hand, ZSL is multi-label imbalanced, i.e. different semantics follow quite different error distributions between classes. As such, existing general imbalanced regression methods are not suitable for the ZSL scenario. For instance, the recent Balanced MSE~\cite{ren2021balanced}, one important general imbalanced learning method,  is  rather limited in dealing with ZSL, which is also empirically demonstrated in our experiments.
	

	To this end, we first statistically  examine the effect of semantic labels on imbalanced semantic prediction.
	Next, leveraging a novel notion of the class-averaged semantic error matrix, we develop a simple yet effective re-balanced strategy, \textbf{Re-balanced Mean-squared Error (ReMSE)} loss, which mitigates the inherited imbalance drawback observed in ZSL.
	To penalize under-fitting semantics potentially with more errors, ReMSE relies on the statistics of semantic errors to generate re-weighting factors.
	Besides, to adjust different error distributions, we also design two-level re-weighting factors, i.e. 1) Class-level re-weighting that compares errors of the same class across different semantics, and 2) Semantic-level re-weighting that compares errors of the same semantic across classes.
	Analytically, we show that minimizing the ReMSE loss tends to minimize the mean error loss as well as the standard deviation of error losses across different semantics, thus certifying the effectiveness of ReMSE in learning a re-balanced visual-semantic mapping.
	In summary, our main contributions are three-fold:
	\begin{enumerate}
		\item[(1)] 
		It is a first attempt to theoretically and empirically analyze that ZSL  is an imbalanced regression problem affected by semantic label values, thus offering a new insight into ZSL.
		\item[(2)] A novel loss function ReMSE is designed for ZSL which dynamically perceives multiple error distributions, focusing on under-fitting semantics without increasing inference cost. Furthermore, We show that minimizing the ReMSE loss tends to  minimize the mean and variance of the error distributions, leading to a rebalanced ZSL.
		\item[(3)] 
		Extensive experiments on three ZSL benchmarks show that our ReMSE effectively alleviates the imbalanced regression problem. Without bells and whistles, our approach outperforms many state-of-the-arts in ZSL (e.g. Transformers), as well as the imbalanced regression: Balanced MSE.
	\end{enumerate}
	
	\section{Related Work}
	\label{sec:rel}
	\subsection{Zero-shot Learning}
	Existing ZSL~\cite{xian2018zero, wei2021lifelong, wei2022incremental, wei2019adversarial} can be mainly divided into generative methods and embedding methods.
	Generative methods utilize generative models, e.g. Generative Adversarial Network~(GAN)~\cite{li2019alleviating,ye2019sr, xie2022leveraging}, Variational AutoEncoder~(VAE)~\cite{li2021generalized}, and Flow models~\cite{shen2020invertible} to synthesize unseen visual features.
	
	In this paper, we focus on the embedding methods~\cite{fu2017zero, chen2022transzero, li2022siamese, wei2021incremental}, which typically learn visual-semantic mapping and classify unseen classes in the mapping space.
	In terms of representation of semantics, word2vec/text2vec~\cite{xian2018zero} leverages pre-trained language models (e.g. Glove~\cite{pennington2014glove}) to provide continuous semantic labels from online text or class names. Manually defining attributes~\cite{lampert2013attribute} is another popular  approach. Vision-based  embedding~\cite{li2018discriminative,ye2021disentangling,xu2022vgse} exploits deep models to  mine latent semantics automatically, which are generally considered more discriminative than attributes.
	
	Most recent embedding proposals study ZSL from the model perspective, i.e. by introducing more complex modules (e.g. GNN)~\cite{fu2017zero} or frameworks (e.g. Transformer)~\cite{chen2022transzero} to extract highly powerful visual features. However, there are just a few investigations of ZSL from the data perspective.
	For example, \cite{akata2015label} shows that $l_{2}$ normalization can compress the noise of semantic labels.
	Most of these works do not explicitly analyze the impact of semantics.
	In contrast, we focus on how semantic annotations would impact learning difficulty. Identifying the semantic prediction imbalance of ZSL, we propose the novel ReMSE algorithm. Supported by the rebalancing strategy, simple CNN-based models can lead to superior performance, even on par with other complex models.
	
	It is worth mentioning that our idea can also benefit generative ZSL methods. For instance, BSeGN~\cite{xie2022leveraging}, one recent generative ZSL method,  applies a different balancing strategy. It generates more realistic visual features through a semantic regressor to classify false visual features. To predict the correlation between samples and classes, BSeGN takes a balancing loss to make closer the classification probabilities of fake samples on seen and unseen classes. Unfortunately, BSeGN does not impose any balance constraints on semantic regressors. Therefore, BSeGN still suffers from the semantic prediction imbalance  problem in the regressor. In contrast, our ReMSE addresses this issue by training a more balanced regressor.
	
	\subsection{Imbalanced Learning}
	
	Despite its long history, imbalanced learning has recently received increasing attention due to its widespread applications in real-world scenarios. This study can be divided into sample distribution-based and forecast error-based methods. Furthermore, most imbalanced learning focuses on classification~\cite{cao2019learning, hu2020learning}, while imbalanced regression involving continuous and infinite target values was first defined in~\cite{yang2021delving}. Below we review some related work on addressing the imbalance problem.
	
	Focal loss~\cite{lin2017focal} is a typical imbalanced prediction error based method for solving binary imbalanced  classification. Focal loss differentiates the difficulty level of samples as a function of the posterior probability $p \in [0,1]$. For the positive class (or negative class), the smaller (larger) $p$ is, the greater the difficulty. Different from focal loss, our loss function is designed to alleviate the imbalance problem of multi-label regression. In addition, the difficulty of predicting a certain class attribute of a sample in our method is closely related to the average regression error $e \in [0,\infty)$ of the attribute. Moreover, the difficulty of prediction is an increasing function of the average regression error.
	
	AdaBoost~\cite{bernhard2013empirical} is another typical method by  adjusting sample weights based on prediction error. It is often used in general multi-class classification tasks, which tries to construct multiple weak classifiers through sample-level reweighting. To reweight each sample, AdaBoost adjusts its weights so that they are positively correlated with the prediction error rate across all samples. In other words, different samples share different weights. In our work, we find that existing ZSL methods suffer from multi-label regression imbalance, i.e., their models have uneven error distributions at both the class level and semantic level. To this end, we introduce class-level and semantic-level balance factors to represent the sample weights, which reinforces that samples of the same class share common weights in the semantic dimension, rather than different weights (as used by AdaBoost).
	
	Current imbalanced regression methods are mainly based on adjusting the sample distribution. Balanced MSE~\cite{ren2021balanced} takes balanced sample distributions to represent the true training and test distributions. Despite the good performance, they merely consider optimizing  imbalanced sample distributions. In fact, we find that the imbalanced values of semantic labels also play an important role in ZSL. Therefore, we propose the ReMSE method, which obtains a rebalanced prediction error distribution across both classes and semantics.
	
	\section{Preliminary}
	The main goal of ZSL is to obtain a classifier that can distinguish visual samples (i.e. images) $\mathcal{X}^{u}$ of unseen classes  $\mathcal{C}^{u}$ from the images $\mathcal{X}^{s}$ of seen classes $\mathcal{C}^{s}$, which only appear in the training set, i.e. $\mathcal{C}^{s} \cap \mathcal{C}^{u} = \emptyset$.
	Since existing methods utilize the class-level semantic labels $\mathcal{S}$ (e.g. attributes or word2vec) to bridge the gap between seen and unseen classes, we define the training set as $\mathcal{D}^{tr}=\{(\mathbf{x}_i, \mathbf{s}_i, y_i)| \mathbf{x}_i \in \mathcal{X}^s, \mathbf{s}_i \in \mathcal{S}, y_i \in \mathcal{C}^{s}\}$, where $\mathbf{x}_i$  and $\mathbf{s}_i$ represent image of the $i$-th  sample and its semantic vector, respectively.
	The number of samples in the training set is denoted by $n_{tr}$.
	Similarly, we can denote the test set by $\mathcal{D}^{te}=\{(\mathbf{x}_i,\mathbf{s}_i,y_i)| \mathbf{x}_i \in \mathcal{X}^u, \mathbf{s}_i \in \mathcal{S}, y_i \in \mathcal{C}^{u} \}$, where the testing samples are from unseen classes in ZSL setting.
	In GZSL setting, the testing samples may be taken from seen classes, in which  $\mathcal{X}^u$ and $\mathcal{C}^{u}$ will be replaced by $\mathcal{X}^u \cup \mathcal{X}^s$ and $\mathcal{C}^u \cup \mathcal{C}^s$, respectively. We denote $d_{s}$ and $d_{v}$ as the dimension of semantic label and visual feature.
	It is worth mentioning that the values of semantic labels vary in different datasets.
	For example, on the dataset CUB~\cite{wah2011caltech}, 312 semantic values range from $0$ to $100$.
	
	
	
	We focus on using Embedding ZSL~\cite{xie2019attentive,liu2021goal}  in our work. Such methods  first use a pre-trained model, such as ResNet, as the backbone for extracting visual feature $\tilde{\mathbf{v}}_{i}$ of the image $\mathbf{x}_i$.
	Then a fully connected network (semantic predictor) with a parameter of $\mathbf{W}\in \mathbb{R}^{d_{s} \times d_{v}}$ solves the ZSL task by semantic prediction:
	\begin{equation}
		f(\tilde{\mathbf{v}}_i) = \mathbf{W} \tilde{\mathbf{v}}_{i}= \tilde{\mathbf{s}}_{i}.
	\end{equation}
	Typically, some recent work~\cite{xu2020attribute,liu2021goal,du2022boosting}  trains embedding ZSL models by leveraging the so-called Semantic Cross-Entropy loss (SCE):
	\begin{equation}
		\label{eq:SCE}
		\mathcal{L}_{SCE} = -\log p_{y_i}(\mathbf{x}_{i}),
	\end{equation}
	\begin{equation}
		\label{eq:CP}
		p_{y_i}(\mathbf{x}_i) = \frac{
			e^{ 
				\tau\cos \theta (\tilde{\mathbf{s}}_i, \mathbf{s}_i) 
			} 
		} {
			\sum_{c \in \mathcal{C}_s }  e^{ 
				\tau\cos \theta (\tilde{\mathbf{s}}_i, \mathbf{s}_c) 
			}
		} ,
	\end{equation}
	in which $p_{y_i}(\mathbf{x}_i)$ is the predicted probability of class $y_i$ for the sample $\mathbf{x}_i$, $\tau$ is a scale hyper-parameter, $\theta (\tilde{\mathbf{s}}_i, \mathbf{s}_c)$ is defined as the angle between semantic predictions $\tilde{\mathbf{s}}_i$ of sample $i$ and semantic labels $\mathbf{s}_c$ of class $c$, i.e. $\cos \theta (\tilde{\mathbf{s}}_i, \mathbf{s}_c) = \tilde{\mathbf{t}}_i^{T} \mathbf{t}_c$, where $\tilde{\mathbf{t}}_i$  and $\mathbf{t}_c$ denote the $l_2$-normalized semantic prediction and class semantic label, i.e.
	\begin{align}
		\tilde{\mathbf{t}}_i = \tilde{\mathbf{s}}_i/\|\tilde{\mathbf{s}}_i\|_2 ,
		\mathbf{t}_c = \mathbf{s}_c/\|\mathbf{s}_c\|_2.
	\end{align}
	Specifically, for ZSL, the test sample $\mathbf{x}_{i} \in \mathcal{X}^u$ can be assigned to the best matching class $c^{\prime}$ from the unseen classes $\mathcal{C}^u$:
	\begin{equation}
		c^{\prime} = \underset{c \in \mathcal{C}^u}{\arg \max} \,\cos\theta (\tilde{\mathbf{s}}_i, \mathbf{s}_c).
	\end{equation}
	In the GZSL setting,  the testing samples may be taken from seen classes, in which  $\mathcal{X}^u$ and $\mathcal{C}^{u}$ will be replaced by $\mathcal{X}^u \cup \mathcal{X}^s$ and $\mathcal{C}^u \cup \mathcal{C}^s$, respectively.
	
	\begin{figure*}[htbp]
		\centering
		\begin{minipage}[t]{0.48\linewidth}
			\centering
			\includegraphics[width=9cm]{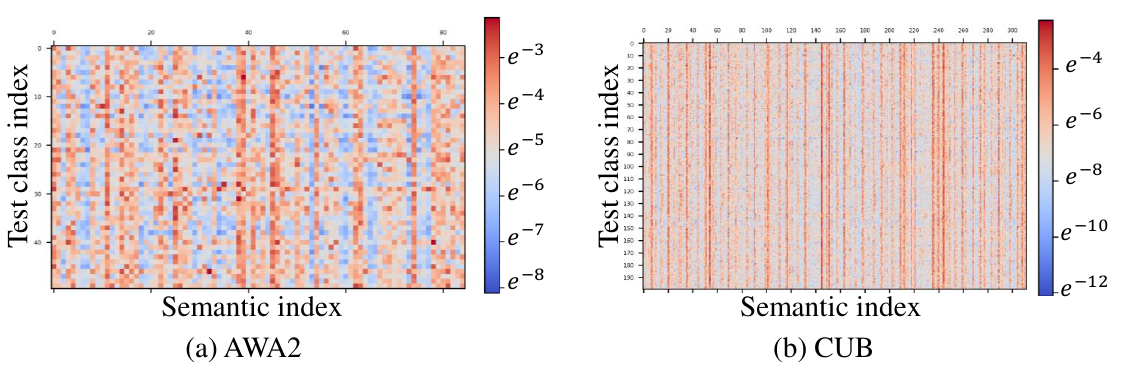}
			\caption{Class-averaged semantic error matrix on test set of AWA2 and CUB under GZSL setting (rescaled by logarithmic function). It shows (1) error distributions are not balanced; (2) the same semantic has different prediction errors on different classes;
				(3) and the same class also has different prediction errors on different semantics.}
			\label{fig:ErrorDistribution}
		\end{minipage}
		\quad
		\begin{minipage}[t]{0.47\linewidth}
			\centering
			\includegraphics[width=8cm]{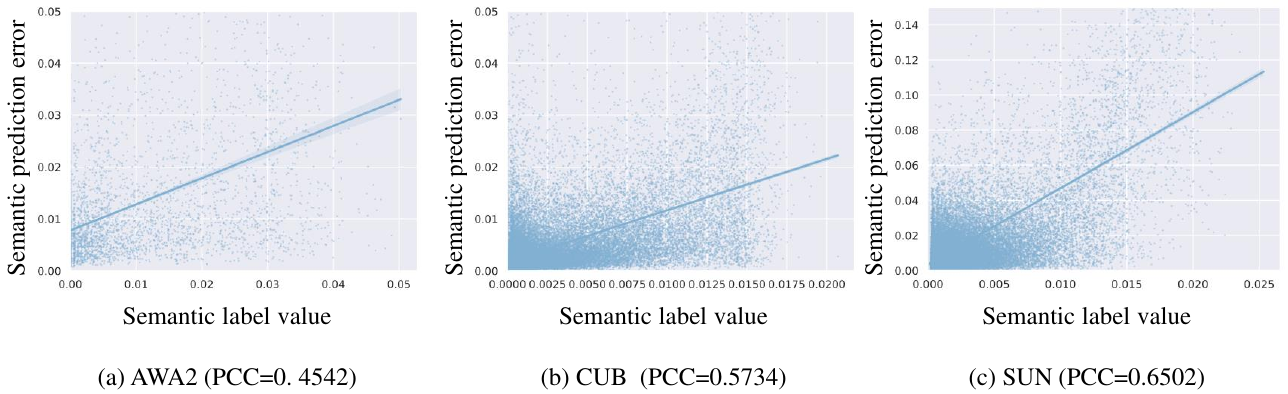}
			\caption{Correlation between semantic label value $t$ and averaged semantic prediction error $m$, which exhibit a strong positive linear relationship regrading Pearson Correlation Coefficient (PCC).}
			\label{fig:correspondence}
		\end{minipage}
	\end{figure*}
	
	\section{Main Methodology}
	
	In this section, we first treat the ZSL as a regression problem. Next, we verify that the imbalanced semantic prediction remains in the existing methods. We then study imbalanced semantic prediction statistically by examining how  semantic label values could affect averaged semantic prediction.
	Finally,  we propose our Rebalanced MSE which could well mitigate the imbalanced semantic prediction issue.
	Our Rebalanced MSE ensures that our model not only minimizes these prediction errors, but also tends to equal or balance these prediction errors whilst converging.
	Furthermore, to enhance the model's representation ability on different semantics, we further design a novel attention-based baseline, named AttentionNet, which can generate semantic-specific attention maps during training and testing stage.

	\subsection{Tackling ZSL via Regression}
	\label{sec:tackling}
	Compared to training  ZSL models with SCE, treating ZSL as a regression task has an unique advantage:
	SCE entangles the gradient of one semantic with other semantics, whereas regression losses (i.e. MSE, MAE and so on) do not.
	
	Specifically, by representing the $k^{th}$ row vector of $\mathbf{W}$ as $\mathbf{w}_{k}$ responsible for  the $k^{th}$ semantic prediction, we could derive the gradient as
	\begin{align}
		\label{eq:SCEgrad}
		\frac{\partial \mathcal{L}_{SCE}}{\partial \mathbf{w}_{k} } & = \sum_{c\in \mathcal{C}^{s}} \frac{\partial \mathcal{L}_{SCE}}{\partial \tilde{\mathbf{t}}_i^{\top} \mathbf{t}_{c} } \frac{\partial \tilde{\mathbf{t}}_i^{\top} \mathbf{t}_{c} } {\partial \mathbf{w}_{k} } \nonumber \\
		& =  \sum_{c \in \mathcal{C}^{s}}  \underbrace{ \left( p_{y_i}(\mathbf{x}_i) - \mathds{1}_{[c=y_i]}  \right) }_{\text{class probability term}}  \frac{\partial \tilde{\mathbf{t}}^{\top} \mathbf{t}_{j} } {\partial \mathbf{w}_{k}}.
	\end{align}
	Obviously, the class probability is determined by the whole semantic prediction. Thus, even if some semantic predictions are not good, the predictions of these semantics can still be further optimized when the class probability is close to 1.
	
	In contrast, regression losses could explicitly perceive the performance of every semantic.
	For example, denoting the formulation of MSE:
	\begin{equation}
		\mathcal{L}_{MSE} =  \frac{1}{N} \sum_{i = 1}^N \|\tilde{\mathbf{s}}_i  - \mathbf{s}_i\|_2^2,
	\end{equation}
	we can obtain the gradient for $\mathbf{w}_{k}$:
	\begin{equation}
		\label{eq:MSE}
		\frac{\partial \mathcal{L}_{MSE}}{\partial \mathbf{w}_{k} } = 2(\mathbf{w}_{k}^\top \tilde{\mathbf{v}}_i - s_{ik}) \tilde{\mathbf{v}}_i.
	\end{equation}
	It is evident that the optimization of $\mathbf{w}_{k}$ would not be interfered by other semantics.
	It is worth mentioning that many works~\cite{romera2015embarrassingly, qiao2016less, xu2020attribute, liu2021goal} tried to utilize the regression loss $\mathcal{L}_{MSE}$ as a compensatory loss in ZSL, i.e.,
	\begin{equation}
		\mathcal{L} = \mathcal{L}_{SCE} + \lambda \mathcal{L}_{MSE},
	\end{equation}
	where $\lambda$ denote a hyper-parameter.
	
	Although $\mathcal{L}_{MSE}$ penalizes the discrepancy between semantic labels and semantic predictions with the Euclidean distance, it is incompatible with the cosine distance optimized by the SCE loss in ZSL.
	We propose a proposition on the normalization of MSE,\footnote{See Appendix A for details.}
	which shows that the original MSE may not precisely measure how well the semantics fit.
	Thus, we normalize both semantic labels and predictions in MSE to obtain the Normalized MSE $\mathcal{L}_{NMSE}(\tilde{\mathbf{s}}_i,\mathbf{s}_i) = \mathcal{L}_{MSE}(\tilde{\mathbf{t}}_i,\mathbf{t}_i)$, which proves fairly compatible with $\mathcal{L}_{SCE}$, since  $\mathcal{L}_{NMSE}= 2- \cos\theta(\tilde{\mathbf{s}}_i, \mathbf{s}_i)$.
	Besides, it is also noted that Balanced MSE~\cite{ren2021balanced} handles imbalanced error distributions by assuming that the distribution of sample is balanced \footnote{See Appendix C for the formulation of Balanced MSE.}.
	However, as discussed in the next section, we find that the imbalanced error distribution in ZSL is caused by the imbalanced semantic values, rather than the sample size. As a result, the Balanced MSE does not perform well in the ZSL setting.
	
	\subsection{Imbalanced Semantic Prediction}
	\label{sec:ISP}
	Now, we consider the imbalanced semantic prediction problem in ZSL.
	During the training process, we find that previous methods often generate imbalanced semantic error distribution.
	To qualitatively illustrate this, in the GZSL setting\footnote{Note that the test samples of unseen classes in ZSL are the same as in GZSL. Thus we only need visualize error matrices under the GZSL setting.}, we exploit GEMZSL (an advanced ZSL method) to visualize the class-averaged semantic error matrix $M \in \mathcal{R}^{|\mathcal{C}^u \cup \mathcal{C}^s| \times d_{s}}$ on AWA2 and CUB.
	Specifically, we collect test samples from every class $l \in \mathcal{C}^u \cup \mathcal{C}^s$ and compute the mean prediction error for each class and each semantic
	\begin{equation} \label{eqn:prediction_loss}
		m_{lj} = \frac{1}{|\mathcal{D}_{l}|} 
		\sum_{(\mathbf{x}_i,\mathbf{s}_i,y_i) \in \mathcal{D}_{l}} (\tilde{t}_{ij}-t_{ij})^2,
	\end{equation}
	where $|\mathcal{D}_l|$ is the number of samples in $\mathcal{D}_l$ and $t_{ij}$ is the $j^{th}$ element of the vector $\mathbf{t}_i$. Notice that $m_{lj}$ indicates the averaged error loss on the $l^{th}$ class for the $j^{th}$ semantic.
	
	The visualization of GZSL on CUB and AWA2 is shown in Fig.~\ref{fig:ErrorDistribution}.
	The results in Fig.~\ref{fig:ErrorDistribution} reveal three interesting observations: (1) ZSL models perform unbalanced on semantic prediction problems: for a certain semantic of a certain class, the model may fit well,  it may however do poorly for other semantics and classes; (2) the same semantic has different prediction errors on different classes; and (3)  the same class also has different prediction errors on different semantics.
	
	To investigate further, in Fig.~\ref{fig:correspondence}, we employ the Pearson Correlation Coefficient (PCC) to quantitatively measure the correlation between the semantic prediction error $m_{lj}$ and semantic label values $t_{ij}$ on three benchmarks.
	It is found that this error loss tends to have a high positive correlation with semantic label value. In other words, the larger the semantic label value, the larger the semantic prediction error.

	
	
	\subsection{ReMSE}
	
	To this end, we try to utilize imbalanced regression methods to balance the semantic errors.
	However, in our experiment, previous imbalanced methods~\cite{ren2021balanced,yang2021delving} (i.e. $\mathcal{L}_{balMSE}$) is not suitable for the imbalance semantic prediction problem.
	They all rely on the hypothesis: imbalanced test error is caused by imbalanced sample distribution, while  in ZSL the imbalanced problem is related to the imbalanced semantic label values, as verified in our work.
	Thus, unlike these methods, we directly adjust the loss, which explicitly measures the semantic prediction performance.
	
	\begin{figure*}[ht]
		\centering
		\includegraphics[width=0.9\linewidth]{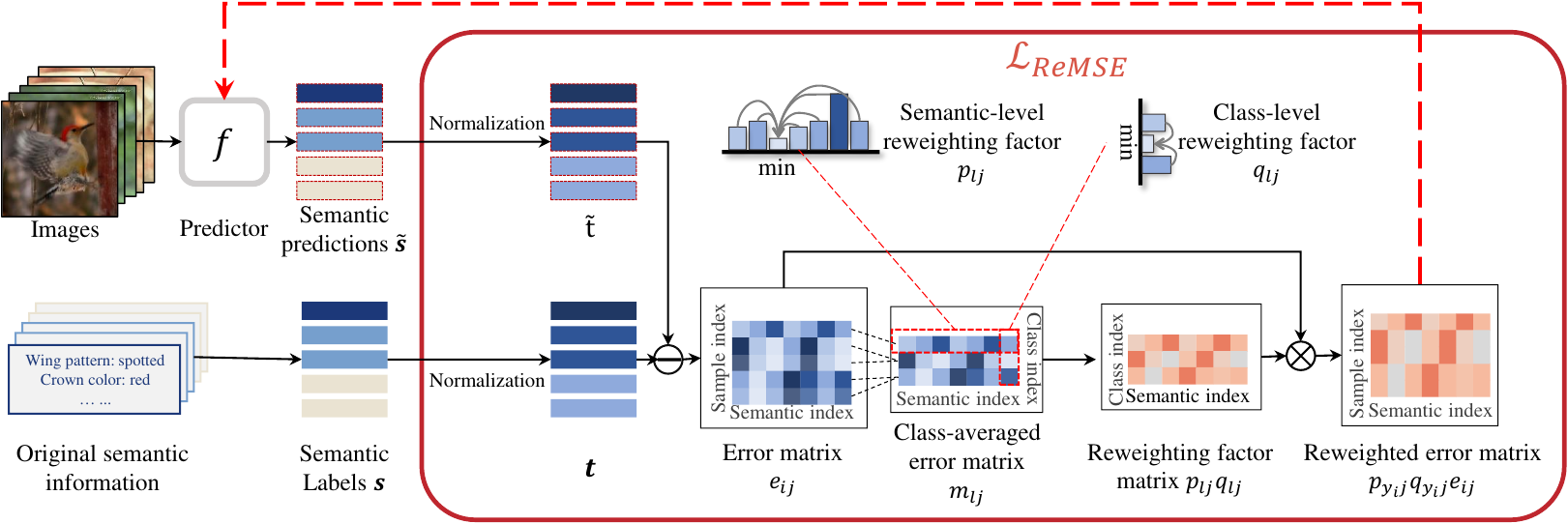}
		\caption{Overview of our ReMSE method: The prediction error is computed on the basis that both the semantic and its predicted value are normalized. Our goal is to minimize the weighted prediction error, where the weight factor of prediction error is represented by the average of prediction errors for the classes, including the category-level factors and the semantic-level factors. It is worth noting that once a new prediction error is obtained, its weight need be recalculated. }
		\label{fig:ReMSE}
	\end{figure*}

	First, we design another class-averaged semantic error matrix $M^{\prime} \in \mathcal{R}^{|\mathcal{C}^s| \times d_{s}}$ within every training batch to establish a common scale for error magnitudes and re-balance every error per class and per semantic.
	As shown in Fig.~\ref{fig:ErrorDistribution}, for different datasets, underfitting of semantic predictions could be distinct: the same semantic has different prediction errors on different classes, and the same class also has different prediction errors on different semantics.
	Thus, given a label $l$ and a semantic $j$, on one hand, we adopt a semantic-level re-weight factor $p_{lj}$ to balance its weight among different semantics of the same class; on the other hand, we adopt a class-level re-weight factor $q_{lj}$ among different classes of the same semantic.
	The class-level balancing factor $p_{lj}$ is designed by
	\begin{equation}
		p_{lj} = \left(\log \frac{ 
			m^{\prime}_{lj} 
		}{
			\min_{c \in \mathcal{C}^s} m^{\prime}_{cj} 
		} + 1 \right)^{\alpha},  \quad  \alpha \ge 0.
	\end{equation}
	A logarithmic function is used here to avoid potential ratio explosions, and a parameter $\alpha$ is taken to control the scale of re-weighting.
	Similarly, the semantic-level balancing factor $q_{lj}$ is calculated as follows:
	\begin{equation}
		q_{lj} = \left(\log \frac{
			m^{\prime}_{lj}
		}{
			\min_{1 \le k \le d_s} m^{\prime}_{l k}
		} + 1 \right)^{\beta},  \quad  \beta \ge 0.
	\end{equation}
	Finally, we obtain the Rebalanced MSE loss function:
	\begin{equation}
		\label{eq:ReMSE}
		\mathcal{L}_{ReMSE}  = \frac{1}{N} \sum^{N}_{i=1} \sum^{d_{s}}_{j=1} p_{y_i j} q_{y_i j} e_{ij},
	\end{equation}
	where $ e_{ij}= (\tilde{t}_{ij}-t_{ij})^2$.
	
	In order to better understand the property of our ReMSE, we reduce our problem to the simplest case, proving that the prediction error for each class will be equal when the algorithm converges\footnote{See Appendix B for the proof of convergence.}.
	
	\subsection{AttentionNet}
	
	\begin{figure}[ht]
		\centering
		\includegraphics[width=0.95\linewidth]{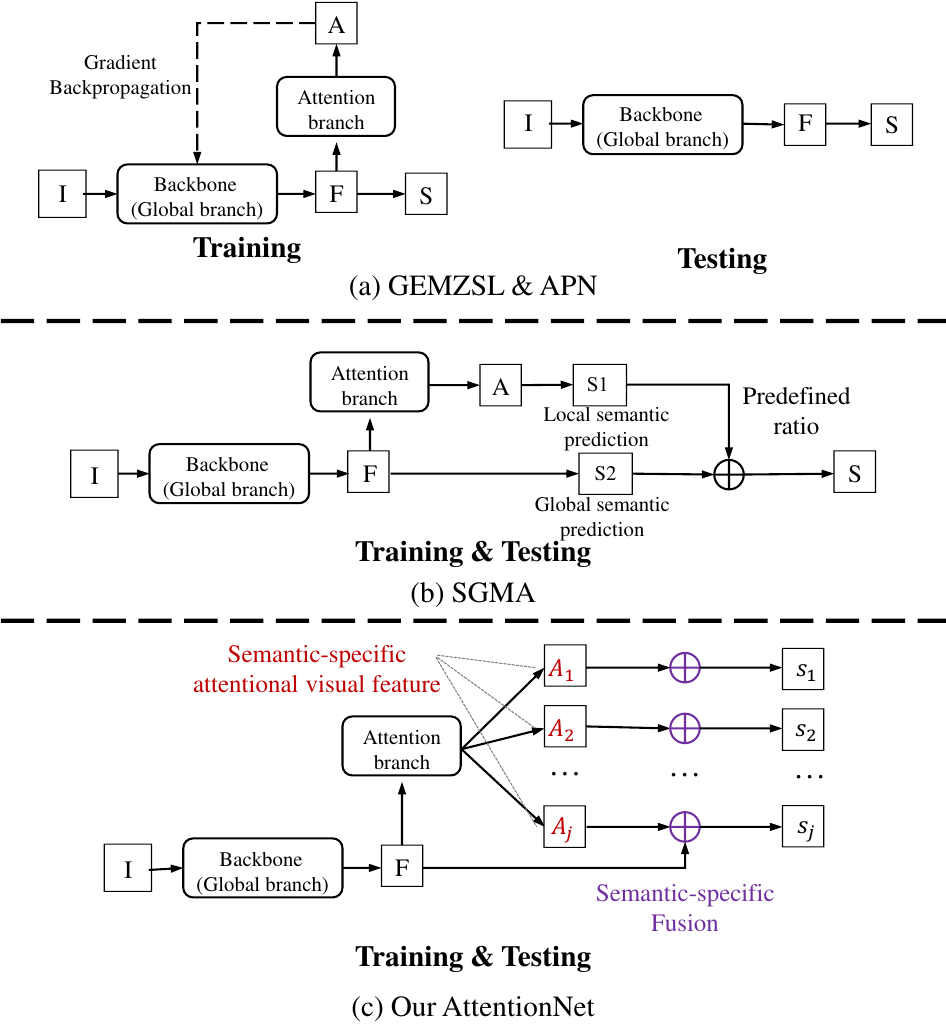}
		\caption{Comparison between previous attention-based ZSL methods and our AttentionNet. The symbols I, F, A and S denote the images, feature maps, attentive features and predicted semantics, respectively. (a) GEMZSL~\cite{liu2021goal} \& APN~\cite{xu2020attribute} use only attention branch to train their backbone, and do not use attention in testing stage. (b) SGAM~\cite{zhu2019semantic} uses attention branch and global branch to predict semantics either locally or globally. (c) In contrast, our AttentionNet utilizes the attention branch in both training and testing stages to produce semantic-specific attentional visual features. Furthermore, we add a semantic-specific feature fusion to avoid the degradation of attentional features.}
		\label{fig:innovation}
	\end{figure}
	
	\begin{figure*}[ht]
		\centering
		\includegraphics[width=0.95\linewidth]{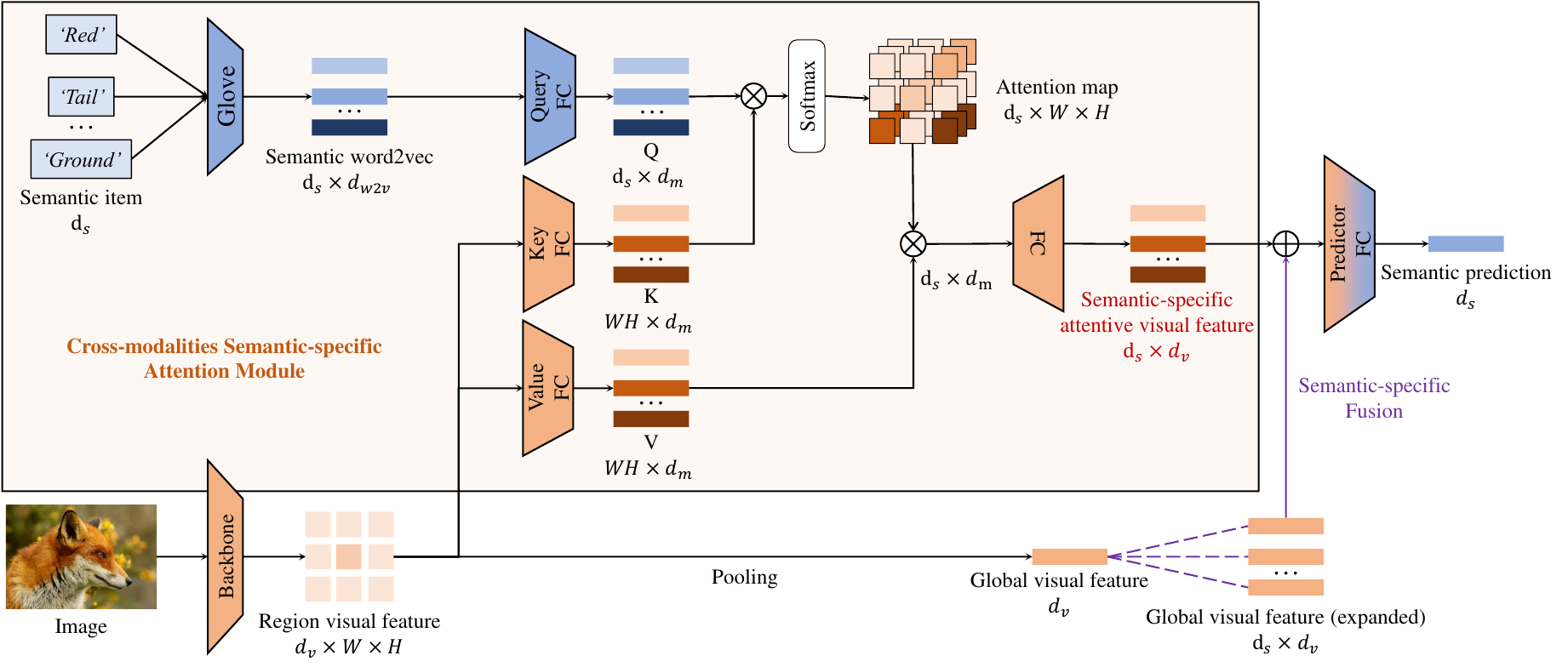}
		\caption{Architecture of our AttentionNet. It contains two innovations: (1) a cross-modal semantic-specific attention module and (2) the semantic-specific fusion. The cross-modal semantic-specific attention module allows the model to pay attention to different local regions according to different semantic attributes to fully exploit both visual and textual semantics. During training and testing, our model can better train end-to-end global and local image features.}
		\label{fig:attentionNet}
	\end{figure*}
	To further improve the accuracy of semantic prediction, we try to utilize additionally the attention mechanism to extract semantic-specific visual features. 
	However, we observe that many existing attention-based ZSL methods do not fully exploit the capacity of the attention mechanism.
	Thus, we design a novel ZSL embedding model called AttentionNet.
	The comparison between existing attention-based ZSL and our AttentionNet is illustrated in Fig.~\ref{fig:innovation}.
	Several remarks are highlighted as follows.
	(1) The traditional methods APN~\cite{xu2020attribute} and GEMZSL~\cite{liu2021goal} only use the attention branch to train the backbone, while in testing stage they abandon it.
	Obviously,  the attention branch cannot help the model to localize discriminative region at testing stage.
	(2) The method SGMA~\cite{zhu2019semantic} utilizes a global branch and an attention branch to extract global and local features and predict semantics individually.
	However, a vast amount of research demonstrates that features fusing local and global information is more powerful~\cite{guo2020augfpn,gao2019res2net}.
	In addition, the predefined ratio also increases the cost of hyper-
	parameter search.
	(3) In contrast, our AttentionNet utilizes the attention branch in both training and testing stages so as to fully exploit the capacity of attention.
	Besides, our attention branch could generate semantic-specific attentive visual features for different semantics. Moreover, to avoid the degradation of attentional features, we add a feature fusion operator between the attentive features and global features, which shows high effectiveness as empirically demonstrated in our Experiment.

	The structure details of our AttentionNet is shown in Fig.~\ref{fig:attentionNet}.
	Specifically, AttentionNet is divided into two branches: the upper branch will synthesize semantic features for image attention, while the lower branch will generate visual (image) global features. In the upper branch, we first adopt the word representation model Glove~\cite{pennington2014glove} to obtain word vectors for all $d_{s}$ semantics ($d_{w2v}$ represents its dimension). For visual feature extraction, we adopt ResNet101~\cite{he2016deep} as our backbone. It will extract $W\times H$ regional visual features using $d_{v}$ channels, which is then followed by our cross-modal attention module. We use three fully connected (FC) layers including query layer, key layer, and value layer to project semantic word2vec and visual features into the latent space where visual and semantic will be aligned. Matrix multiplication between $Q$ and $K$ represents the strength of attention (normalized by softmax). Once the attention map is obtained, the attention network will drive the semantic items to automatically focus on specific regions. Next, our network augments the vector $d_v$ to obtain $d_s \times d_v$ features, which are then merged with the upper branch as the input of the semantic prediction network.
	
	Our cross-modal semantic-specific attention module between semantics and images allows our model to implicitly establish relationships between semantic attributes and image features, which could take full advantages of visual and textual modalities.
	With the help of the attention mechanism, our model is able to obtain better local image features, which are related to the predicted semantics, and fused with global features in the testing phase to obtain better semantic predictions.

	\section{Experiments}
	\begin{table*}[htbp]
		\centering
		\caption{Statistics of datasets.}
		\label{table:dataset}
		\begin{tabular}{c|c|c|c|c|c|c|c}
			\hline
			Dataset&\tabincell{c}{Semantic\\dimension} &\tabincell{c}{Semantic\\range}&\tabincell{c}{$\#$ Seen\\classes}&\tabincell{c}{$\#$ Unseen\\classes}&\tabincell{c}{$\#$ Images\\(total)}&\tabincell{c}{$\#$ Images\\(train+val)}&\tabincell{c}{$\#$ Images\\(test unseen/seen)}\\
			\hline
			AWA2~\cite{lampert2013attribute}&85 & $[0,100] \cup \{-1\}$ &40&10&30475&19832&4958/5685\\
			\hline
			CUB~\cite{wah2011caltech}&312& $[0,100]$&150&50&11788&7057&2679/1764\\
			\hline
			SUN~\cite{patterson2012sun}&102& $[0,1]$ &645&72&14340&10320&1440/2580\\
			\hline
		\end{tabular}
	\end{table*}
	
	\begin{table*}[htb]
		\centering
		\caption{Overall comparison with SOTAs in the setting of ZSL and GZSL.
			In ZSL, T1 represents  the top-1 accuracy (\%)  for unseen classes.
			In  GZSL, $U$, $S$ and $H$ represent the top-1 accuracy (\%) of unseen classes, seen classes, and their harmonic mean, respectively.
			W2V indicates whether the methods use semantic-level word2vec.
			The symbol $^\star$ represents the results of our implemented version.  The best and second best results in the embedding methods are marked with \textcolor{red}{\textbf{red}} and \textcolor{blue}{\textbf{blue}}. }
		\begin{tabular}{p{2.3cm}|p{0.2cm}|p{1.2cm}|p{0.05cm}p{0.05cm}p{0.05cm}|p{0.05cm}p{0.05cm}p{0.05cm}p{0.05cm}p{0.05cm}p{0.05cm}p{0.05cm}p{0.05cm}p{0.05cm}}
			\hline
			& & &\multicolumn{3}{|c}{Zero-shot Learning} & \multicolumn{9}{|c}{Generalized Zero-shot Learning}\\
			\hline
			& & & \multicolumn{1}{|c}{AWA2} & \multicolumn{1}{|c}{CUB} & \multicolumn{1}{|c|}{SUN} & \multicolumn{3}{|c}{AWA2} & \multicolumn{3}{|c}{CUB} &\multicolumn{3}{|c}{SUN} \\
			\hline
			Approach & \multicolumn{1}{c|}{W2V} & Refer & \multicolumn{1}{|c}{T1} & \multicolumn{1}{|c}{T1} & \multicolumn{1}{|c|}{T1} & \multicolumn{1}{|c}{U} & \multicolumn{1}{c}{S} & \multicolumn{1}{c|}{H} & \multicolumn{1}{|c}{U} & \multicolumn{1}{c}{S} & \multicolumn{1}{c|}{H} & \multicolumn{1}{|c}{U} & \multicolumn{1}{c}{S} & \multicolumn{1}{c}{H} \\
			\hline
			\multicolumn{15}{c}{Embedding approaches}\\
			\hline
			AGEN~\cite{xie2019attentive} & \multicolumn{1}{c|}{$\times$} & CVPR19 & \multicolumn{1}{|c|}{66.9} & \multicolumn{1}{|c|}{72.5} & \multicolumn{1}{|c|}{60.6} & \multicolumn{1}{|c}{54.7} & \multicolumn{1}{c}{79.1} & \multicolumn{1}{c|}{64.7} & \multicolumn{1}{|c}{63.2} & \multicolumn{1}{c}{69.0} & \multicolumn{1}{c|}{66.0} & \multicolumn{1}{|c}{40.3} & \multicolumn{1}{c}{32.3} & \multicolumn{1}{c}{35.9}\\
			DUET~\cite{jia2019deep} & \multicolumn{1}{c|}{$\times$} & TIP19 & \multicolumn{1}{|c|}{\textcolor{red}{\textbf{72.6}}} & \multicolumn{1}{|c|}{72.4} & \multicolumn{1}{|c|}{-} & \multicolumn{1}{|c}{48.2} & \multicolumn{1}{c}{\textcolor{red}{\textbf{90.2}}} & \multicolumn{1}{c|}{63.4} & \multicolumn{1}{|c}{39.7} & \multicolumn{1}{c}{80.1} & \multicolumn{1}{c|}{53.1} & \multicolumn{1}{|c}{-} & \multicolumn{1}{c}{-} & \multicolumn{1}{c}{-} \\
			SGAM~\cite{zhu2019semantic} & \multicolumn{1}{c|}{$\times$} & NeurIPS19 & \multicolumn{1}{|c|}{68.8} & \multicolumn{1}{|c|}{71.0} & \multicolumn{1}{|c|}{-} & \multicolumn{1}{|c}{37.6} & \multicolumn{1}{c}{\textcolor{blue}{\textbf{87.1}}} & \multicolumn{1}{c|}{52.5} & \multicolumn{1}{|c}{36.7} & \multicolumn{1}{c}{71.3} & \multicolumn{1}{c|}{48.5} & \multicolumn{1}{|c}{-} & \multicolumn{1}{c}{-} & \multicolumn{1}{c}{-} \\
			APN~\cite{xu2020attribute} & \multicolumn{1}{c|}{$\times$} & NeurIPS20 & \multicolumn{1}{|c|}{68.4} & \multicolumn{1}{|c|}{72.0} & \multicolumn{1}{|c|}{61.6} & \multicolumn{1}{|c}{56.5} & \multicolumn{1}{c}{78.0} & \multicolumn{1}{c|}{65.5} & \multicolumn{1}{|c}{65.3} & \multicolumn{1}{c}{69.3} & \multicolumn{1}{c|}{67.2} & \multicolumn{1}{|c}{41.9} & \multicolumn{1}{c}{34.0} & \multicolumn{1}{c}{37.6}\\
			GEMZSL~\cite{liu2021goal}  & \multicolumn{1}{c|}{$\checkmark$} & CVPR21 & \multicolumn{1}{|c|}{67.3} & \multicolumn{1}{|c|}{77.8} & \multicolumn{1}{|c|}{62.8} & \multicolumn{1}{|c}{64.8} & \multicolumn{1}{c}{77.5} & \multicolumn{1}{c|}{70.6} & \multicolumn{1}{|c}{64.8} & \multicolumn{1}{c}{\textcolor{red}{\textbf{77.1}}} & \multicolumn{1}{c|}{70.4} & \multicolumn{1}{|c}{38.1} & \multicolumn{1}{c}{35.7} & \multicolumn{1}{c}{36.9} \\
			LSG~\cite{xu2021semi} & \multicolumn{1}{c|}{$\times$} & TIP21 & \multicolumn{1}{|c|}{61.1} & \multicolumn{1}{|c|}{52.9} & \multicolumn{1}{|c|}{53.4} & \multicolumn{1}{|c}{60.4} & \multicolumn{1}{c}{84.9} & \multicolumn{1}{c|}{70.6} & \multicolumn{1}{|c}{49.6} & \multicolumn{1}{c}{50.4} & \multicolumn{1}{c|}{50.0} & \multicolumn{1}{|c}{\textcolor{red}{\textbf{52.8}}} & \multicolumn{1}{c}{23.1} & \multicolumn{1}{c}{32.2} \\
			TransZero~\cite{chen2022transzero}  & \multicolumn{1}{c|}{$\checkmark$} & AAAI22 & \multicolumn{1}{|c|}{70.1} & \multicolumn{1}{|c|}{76.8} & \multicolumn{1}{|c|}{65.6} & \multicolumn{1}{|c}{61.3} & \multicolumn{1}{c}{82.3} & \multicolumn{1}{c|}{70.2} & \multicolumn{1}{|c}{69.3} & \multicolumn{1}{c}{68.3} & \multicolumn{1}{c|}{68.8} & \multicolumn{1}{|c}{\textcolor{blue}{\textbf{52.6}}} & \multicolumn{1}{c}{33.4} & \multicolumn{1}{c}{\textcolor{blue}{\textbf{40.8}}} \\
			TransZero++~\cite{chen2022transzero++}  & \multicolumn{1}{c|}{$\checkmark$} & TPAMI22 & \multicolumn{1}{|c|}{\textcolor{red}{\textbf{72.6}}} & \multicolumn{1}{|c|}{78.3} & \multicolumn{1}{|c|}{\textcolor{red}{\textbf{67.6}}} & \multicolumn{1}{|c}{\textcolor{red}{\textbf{64.6}}} & \multicolumn{1}{c}{82.7} & \multicolumn{1}{c|}{72.5} & \multicolumn{1}{|c}{67.5} & \multicolumn{1}{c}{73.6} & \multicolumn{1}{c|}{70.4} & \multicolumn{1}{|c}{48.6} & \multicolumn{1}{c}{\textcolor{blue}{\textbf{37.8}}} & \multicolumn{1}{c}{\textcolor{red}{\textbf{42.5}}} \\
			\hline
			APN$^\star$~\cite{xu2020attribute} & \multicolumn{1}{c|}{$\times$} & NeurIPS20 & \multicolumn{1}{|c|}{68.2} & \multicolumn{1}{|c|}{71.9} & \multicolumn{1}{|c|}{61.0} & \multicolumn{1}{|c}{59.8} & \multicolumn{1}{c}{75.1} & \multicolumn{1}{c|}{66.6} & \multicolumn{1}{|c}{64.4} & \multicolumn{1}{c}{67.8} & \multicolumn{1}{c|}{66.0} & \multicolumn{1}{|c}{41.1} & \multicolumn{1}{c}{34.0} & \multicolumn{1}{c}{37.2} \\
			~~~~+Balanced MSE & \multicolumn{1}{c|}{$\times$} & CVPR22 &\multicolumn{1}{|c|}{68.1} & \multicolumn{1}{|c|}{68.5} & \multicolumn{1}{|c|}{60.6} & \multicolumn{1}{|c}{58.3} & \multicolumn{1}{c}{78.9} & \multicolumn{1}{c|}{67.1} & \multicolumn{1}{|c}{57.0} & \multicolumn{1}{c}{65.5} & \multicolumn{1}{c|}{60.9} & \multicolumn{1}{|c}{41.3} & \multicolumn{1}{c}{34.3} & \multicolumn{1}{c}{37.4} \\
			~~~~\textbf{+ReMSE} & \multicolumn{1}{c|}{$\times$} & \textbf{Ours} &\multicolumn{1}{|c|}{68.3} & \multicolumn{1}{|c|}{72.1} & \multicolumn{1}{|c|}{61.5} & \multicolumn{1}{|c}{63.2} & \multicolumn{1}{c}{74.9} & \multicolumn{1}{c|}{68.5} & \multicolumn{1}{|c}{67.8} & \multicolumn{1}{c}{64.7} & \multicolumn{1}{c|}{66.2} & \multicolumn{1}{|c}{42.9} & \multicolumn{1}{c}{33.7} & \multicolumn{1}{c}{37.7} \\
			\hline
			GEMZSL$^\star$~\cite{liu2021goal}  & \multicolumn{1}{c|}{$\checkmark$} & CVPR21 &\multicolumn{1}{|c|}{65.7} & \multicolumn{1}{|c|}{75.8} & \multicolumn{1}{|c|}{62.2} & \multicolumn{1}{|c}{62.0} & \multicolumn{1}{c}{79.9} & \multicolumn{1}{c|}{69.8} & \multicolumn{1}{|c}{69.9} & \multicolumn{1}{c}{73.2} & \multicolumn{1}{c|}{71.5} & \multicolumn{1}{|c}{37.3} & \multicolumn{1}{c}{\textcolor{red}{\textbf{37.9}}} & \multicolumn{1}{c}{37.6} \\
			~~~~+Balanced MSE  & \multicolumn{1}{c|}{$\checkmark$} & CVPR22 &\multicolumn{1}{|c|}{65.3} & \multicolumn{1}{|c|}{75.3} & \multicolumn{1}{|c|}{61.7} & \multicolumn{1}{|c}{60.7} & \multicolumn{1}{c}{81.2} & \multicolumn{1}{c|}{69.5} & \multicolumn{1}{|c}{67.1} & \multicolumn{1}{c}{\textcolor{blue}{\textbf{75.5}}} & \multicolumn{1}{c|}{71.1} & \multicolumn{1}{|c}{46.3} & \multicolumn{1}{c}{30.9} & \multicolumn{1}{c}{37.1} \\
			~~~~\textbf{+ReMSE}  & \multicolumn{1}{c|}{$\checkmark$} & \textbf{Ours} & \multicolumn{1}{|c|}{66.1} & \multicolumn{1}{|c|}{76.6} & \multicolumn{1}{|c|}{63.1} & \multicolumn{1}{|c}{61.4} & \multicolumn{1}{c}{81.9} & \multicolumn{1}{c|}{70.2} & \multicolumn{1}{|c}{69.0} & \multicolumn{1}{c}{75.2} & \multicolumn{1}{c|}{72.0} & \multicolumn{1}{|c}{48.8} & \multicolumn{1}{c}{33.6} & \multicolumn{1}{c}{39.8} \\
			\hline
			\hline
			\textbf{AttentionNet} & \multicolumn{1}{c|}{$\checkmark$} & \textbf{Ours} & \multicolumn{1}{|c|}{69.3} & \multicolumn{1}{|c|}{\textcolor{blue}{\textbf{80.2}}} & \multicolumn{1}{|c|}{62.8} & \multicolumn{1}{|c}{\textcolor{blue}{\textbf{63.8}}} & \multicolumn{1}{c}{84.6} & \multicolumn{1}{c|}{\textcolor{blue}{\textbf{72.8}}} & \multicolumn{1}{c}{\textcolor{blue}{\textbf{71.9}}} & \multicolumn{1}{c}{74.6} & \multicolumn{1}{c|}{\textcolor{blue}{\textbf{72.9}}} & \multicolumn{1}{|c}{47.1} & \multicolumn{1}{c}{32.8} & \multicolumn{1}{c}{38.6} \\
			~~~~+Balanced MSE  & \multicolumn{1}{c|}{$\checkmark$} & CVPR22 & \multicolumn{1}{|c|}{66.4} & \multicolumn{1}{|c|}{79.6} & \multicolumn{1}{|c|}{62.4} & \multicolumn{1}{|c}{59.9} & \multicolumn{1}{c}{83.4} & \multicolumn{1}{c|}{69.7} & \multicolumn{1}{c}{70.7} & \multicolumn{1}{c}{74.9} & \multicolumn{1}{c|}{72.7} & \multicolumn{1}{|c}{47.9} & \multicolumn{1}{c}{33.7} & \multicolumn{1}{c}{39.6} \\
			~~~~\textbf{+ReMSE}  & \multicolumn{1}{c|}{$\checkmark$} & \textbf{Ours} & \multicolumn{1}{|c|}{\textcolor{blue}{\textbf{70.9}}} & \multicolumn{1}{|c|}{\textcolor{red}{\textbf{80.9}}} & \multicolumn{1}{|c|}{63.2} & \multicolumn{1}{|c}{\textcolor{blue}{\textbf{63.8}}} & \multicolumn{1}{c}{85.6} & \multicolumn{1}{c|}{\textcolor{red}{\textbf{73.1}}} & \multicolumn{1}{c}{\textcolor{red}{\textbf{72.8}}} & \multicolumn{1}{c}{74.8} & \multicolumn{1}{c|}{\textcolor{red}{\textbf{73.8}}} & \multicolumn{1}{|c}{47.4} & \multicolumn{1}{c}{34.8} & \multicolumn{1}{c}{40.1} \\
			\hline
			\textbf{AttentionNet (GB)} & \multicolumn{1}{c|}{$\times$} & \textbf{Ours} & \multicolumn{1}{|c|}{67.0} & \multicolumn{1}{|c|}{74.8} & \multicolumn{1}{|c|}{62.7} & \multicolumn{1}{|c}{62.1} & \multicolumn{1}{c}{83.1} & \multicolumn{1}{c|}{71.1} & \multicolumn{1}{|c}{66.9} & \multicolumn{1}{c}{72.0} & \multicolumn{1}{c|}{69.3} & \multicolumn{1}{|c}{47.5} & \multicolumn{1}{c}{31.2} & \multicolumn{1}{c}{37.6}\\
			~~~~\textbf{+ReMSE} & \multicolumn{1}{c|}{$\times$} & \textbf{Ours} & \multicolumn{1}{|c|}{68.8} & \multicolumn{1}{|c|}{77.6} & \multicolumn{1}{|c|}{\textcolor{blue}{\textbf{63.6}}} & \multicolumn{1}{|c}{62.9} & \multicolumn{1}{c}{84.6} & \multicolumn{1}{c|}{72.2} & \multicolumn{1}{|c}{70.5} & \multicolumn{1}{c}{72,7} & \multicolumn{1}{c|}{70.6} & \multicolumn{1}{|c}{48.8} & \multicolumn{1}{c}{33.3} & \multicolumn{1}{c}{39.6}\\
			\hline
			\rowcolor{gray!10}
			\multicolumn{15}{c}{Generative approaches}\\
			\hline
			\rowcolor{gray!10}
			fCLSWGAN~\cite{xian2018feature}  & \multicolumn{1}{c|}{$\times$} & CVPR18 & \multicolumn{1}{|c|}{-} & \multicolumn{1}{|c|}{57.3} & \multicolumn{1}{|c|}{60.8} & \multicolumn{1}{|c}{56.1} & \multicolumn{1}{c}{65.5} & \multicolumn{1}{c|}{60.4} & \multicolumn{1}{|c}{43.7} & \multicolumn{1}{c}{57.7} & \multicolumn{1}{c|}{49.7} & \multicolumn{1}{|c}{42.6} & \multicolumn{1}{c}{36.6} & \multicolumn{1}{c}{39.4}\\
			\rowcolor{gray!10}
			DCRGAN~\cite{ye2021disentangling} & \multicolumn{1}{c|}{$\times$} & TMM21 & \multicolumn{1}{|c|}{-} & \multicolumn{1}{|c|}{61.0} & \multicolumn{1}{|c|}{63.7} & \multicolumn{1}{|c}{-} & \multicolumn{1}{c}{-} & \multicolumn{1}{c|}{-} & \multicolumn{1}{|c}{55.8} & \multicolumn{1}{c}{66.8} & \multicolumn{1}{c|}{60.8} & \multicolumn{1}{|c}{47.1} & \multicolumn{1}{c}{38.5} & \multicolumn{1}{c}{42.4} \\
			\rowcolor{gray!10}
			DisVAE~\cite{li2021generalized} & \multicolumn{1}{c|}{$\times$} & AAAI21& \multicolumn{1}{|c|}{-} & \multicolumn{1}{|c|}{-} & \multicolumn{1}{|c|}{-} & \multicolumn{1}{|c}{56.9} & \multicolumn{1}{c}{80.2} & \multicolumn{1}{c|}{66.6} & \multicolumn{1}{|c}{51.1} & \multicolumn{1}{c}{58.2} & \multicolumn{1}{c|}{54.4} & \multicolumn{1}{|c}{36.6} & \multicolumn{1}{c}{47.6} & \multicolumn{1}{c}{41.4} \\
			\rowcolor{gray!10}
			HSVA~\cite{chen2021hsva} & \multicolumn{1}{c|}{$\times$} & NeurIPS21 & \multicolumn{1}{|c|}{-} & \multicolumn{1}{|c|}{62.8} & \multicolumn{1}{|c|}{63.8} & \multicolumn{1}{|c}{56.7} & \multicolumn{1}{c}{79.8} & \multicolumn{1}{c|}{66.3} & \multicolumn{1}{|c}{52.7} & \multicolumn{1}{c}{58.3} & \multicolumn{1}{c|}{55.3} & \multicolumn{1}{|c}{48.6} & \multicolumn{1}{c}{39.0} & \multicolumn{1}{c}{43.3} \\
			\rowcolor{gray!10}
			CE-GZSL~\cite{han2021contrastive} & \multicolumn{1}{c|}{$\times$} & CVPR21& \multicolumn{1}{|c|}{70.4} & \multicolumn{1}{|c|}{77.5} & \multicolumn{1}{|c|}{63.3} & \multicolumn{1}{|c}{63.1} & \multicolumn{1}{c}{78.6} & \multicolumn{1}{c|}{70.0} & \multicolumn{1}{|c}{63.9} & \multicolumn{1}{c}{66.8} & \multicolumn{1}{c|}{65.3} & \multicolumn{1}{|c}{48.8} & \multicolumn{1}{c}{38.6} & \multicolumn{1}{c}{43.1} \\
			\rowcolor{gray!10}
			BSeGN~\cite{xie2022leveraging} & \multicolumn{1}{c|}{$\times$} & TNNLS22& \multicolumn{1}{|c|}{71.5} & \multicolumn{1}{|c|}{65.3} & \multicolumn{1}{|c|}{66.4} & \multicolumn{1}{|c}{59.3} & \multicolumn{1}{c}{78.0} & \multicolumn{1}{c|}{67.4} & \multicolumn{1}{|c}{55.3} & \multicolumn{1}{c}{60.8} & \multicolumn{1}{c|}{58.0} & \multicolumn{1}{|c}{48.9} & \multicolumn{1}{c}{38.3} & \multicolumn{1}{c}{42.9} \\
			\hline
		\end{tabular}
		\label{table:OverallComparison}
	\end{table*}
	
	\begin{table}[htbp]
		\centering
		\caption{AUSUC for GZSL. A larger value means a better trade-off between seen and unseen accuracy.}
		\begin{tabular}{p{0.1cm}|p{0.1cm}|p{0.1cm}|p{0.1cm}|}
			\hline
			& \multicolumn{1}{c|}{AWA2} & \multicolumn{1}{|c|}{CUB} & \multicolumn{1}{|c}{SUN}\\
			\hline
			\multicolumn{1}{c|}{APN} & \multicolumn{1}{|c|}{0.5784} &  \multicolumn{1}{|c|}{0.5545} & \multicolumn{1}{|c}{0.2056}\\
			\multicolumn{1}{c|}{+Balanced MSE} & \multicolumn{1}{|c|}{0.5825} &  \multicolumn{1}{|c|}{0.4973} & \multicolumn{1}{|c}{0.2064}\\
			\multicolumn{1}{c|}{\textbf{+ReMSE (Ours)}} & \multicolumn{1}{|c|}{\textbf{0.5840}} &  \multicolumn{1}{|c|}{\textbf{0.5552}} & \multicolumn{1}{|c}{\textbf{0.2114}}\\
			\hline
			\multicolumn{1}{c|}{GEMZSL} & \multicolumn{1}{|c|}{0.5823} &  \multicolumn{1}{|c|}{0.6178} & \multicolumn{1}{|c}{0.2211}\\
			\multicolumn{1}{c|}{+Balanced MSE} & \multicolumn{1}{|c|}{0.6017} &  \multicolumn{1}{|c|}{0.6137} & \multicolumn{1}{|c}{0.1991}\\
			\multicolumn{1}{c|}{\textbf{+ReMSE (Ours)}} & \multicolumn{1}{|c|}{\textbf{0.6067}} &  \multicolumn{1}{|c|}{\textbf{0.6230}} & \multicolumn{1}{|c}{\textbf{0.2275}}\\
			\hline
			\multicolumn{1}{c|}{{AttentionNet}} & \multicolumn{1}{|c|}{0.6275} &  \multicolumn{1}{|c|}{0.6390} & \multicolumn{1}{|c}{0.2145}\\
			\multicolumn{1}{c|}{+Balanced MSE} & \multicolumn{1}{|c|}{0.6007} &  \multicolumn{1}{|c|}{0.6373} & \multicolumn{1}{|c}{0.2192}\\
			\multicolumn{1}{c|}{\textbf{+ReMSE (Ours)}} & \multicolumn{1}{|c|}{\textbf{0.6476}} &  \multicolumn{1}{|c|}{\textbf{0.6516}} & \multicolumn{1}{|c}{\textbf{0.2338}}\\
			\hline
		\end{tabular}
		\label{table:AUSUC}
	\end{table}

	\begin{table*}[htbp]
		\centering
		\caption{
			Ablation results in the ZSL and GZSL settings. AB and GB denote the attention branch and the global branch, respectively. In ZSL, T1 represents  the top-1 accuracy (\%)  for unseen classes. In  GZSL, $U$, $S$ and $H$ denote the top-1 accuracy (\%) for unseen classes, seen classes, and their harmonic mean, respectively.
		}
		\begin{tabular}{p{1.5cm}|p{1.0cm}|p{2.0cm}|p{0.04cm}p{0.04cm}p{0.04cm}|p{0.04cm}p{0.04cm}p{0.04cm}p{0.04cm}p{0.04cm}p{0.04cm}p{0.04cm}p{0.04cm}p{0.04cm}}
			\hline
			& & &\multicolumn{3}{|c}{Zero-shot Learning} & \multicolumn{9}{|c}{Generalized Zero-shot Learning}\\
			\hline
			& & & \multicolumn{1}{|c}{AWA2} & \multicolumn{1}{|c}{CUB} & \multicolumn{1}{|c|}{SUN} & \multicolumn{3}{|c}{AWA2} & \multicolumn{3}{|c}{CUB} &\multicolumn{3}{|c}{SUN} \\
			\hline
			Structure & Backbone & Loss & \multicolumn{1}{|c}{T1} & \multicolumn{1}{|c}{T1} & \multicolumn{1}{|c|}{T1} & \multicolumn{1}{|c}{U} & \multicolumn{1}{c}{S} & \multicolumn{1}{c|}{H} & \multicolumn{1}{|c}{U} & \multicolumn{1}{c}{S} & \multicolumn{1}{c|}{H} & \multicolumn{1}{|c}{U} & \multicolumn{1}{c}{S} & \multicolumn{1}{c}{H} \\
			\hline
			GB & ResNet & $\mathcal{L}_{SCE}$ & \multicolumn{1}{|c|}{67.0} & \multicolumn{1}{|c|}{74.8} & \multicolumn{1}{|c|}{62.7} & \multicolumn{1}{|c}{62.1} & \multicolumn{1}{c}{83.1} & \multicolumn{1}{c|}{71.1} & \multicolumn{1}{|c}{66.9} & \multicolumn{1}{c}{72.0} & \multicolumn{1}{c|}{69.3} & \multicolumn{1}{|c}{47.5} & \multicolumn{1}{c}{31.2} & \multicolumn{1}{c}{37.6}\\
			GB & ResNet & $\mathcal{L}_{SCE}$+$\mathcal{L}_{ReMSE}$ & \multicolumn{1}{|c|}{68.8} & \multicolumn{1}{|c|}{77.6} & \multicolumn{1}{|c|}{63.6} & \multicolumn{1}{|c}{62.9} & \multicolumn{1}{c}{84.6} & \multicolumn{1}{c|}{72.2} & \multicolumn{1}{|c}{70.5} & \multicolumn{1}{c}{72,7} & \multicolumn{1}{c|}{70.6} & \multicolumn{1}{|c}{48.8} & \multicolumn{1}{c}{33.3} & \multicolumn{1}{c}{39.6}\\
			AB & ResNet & $\mathcal{L}_{SCE}$ & \multicolumn{1}{|c|}{64.2} & \multicolumn{1}{|c|}{79.7} & \multicolumn{1}{|c|}{60.9} & \multicolumn{1}{|c}{58.3} & \multicolumn{1}{c}{68.5} & \multicolumn{1}{c|}{68.4} & \multicolumn{1}{|c}{71.5} & \multicolumn{1}{c}{73.9} & \multicolumn{1}{c|}{72.7} & \multicolumn{1}{|c}{46.7} & \multicolumn{1}{c}{30.6} & \multicolumn{1}{c}{37.0}\\
			AttentionNet & ResNet & $\mathcal{L}_{SCE}$ & \multicolumn{1}{|c|}{69.3} & \multicolumn{1}{|c|}{80.2} & \multicolumn{1}{|c|}{62.8} & \multicolumn{1}{|c}{63.8} & \multicolumn{1}{c}{84.6} & \multicolumn{1}{c|}{72.8} & \multicolumn{1}{|c}{71.2} & \multicolumn{1}{c}{74.6} & \multicolumn{1}{c|}{72.9} & \multicolumn{1}{|c}{47.1} & \multicolumn{1}{c}{32.8} & \multicolumn{1}{c}{38.6}\\
			AttentionNet & ResNet & $\mathcal{L}_{SCE}$+$\mathcal{L}_{ReMSE}$ & \multicolumn{1}{|c|}{70.9} & \multicolumn{1}{|c|}{80.9} & \multicolumn{1}{|c|}{63.2} & \multicolumn{1}{|c}{63.8} & \multicolumn{1}{c}{85.6} & \multicolumn{1}{c|}{73.1} & \multicolumn{1}{|c}{72.8} & \multicolumn{1}{c}{74.8} & \multicolumn{1}{c|}{73.8} & \multicolumn{1}{|c}{47.4} & \multicolumn{1}{c}{34.8} & \multicolumn{1}{c}{40.1}\\
			\hline
			AttentionNet & ViT & $\mathcal{L}_{SCE}$ & \multicolumn{1}{|c|}{67.1} & \multicolumn{1}{|c|}{76.6} & \multicolumn{1}{|c|}{68.9} & \multicolumn{1}{|c}{60.4} & \multicolumn{1}{c}{83.6} & \multicolumn{1}{c|}{70.2} & \multicolumn{1}{|c}{69.9} & \multicolumn{1}{c}{75.2} & \multicolumn{1}{c|}{72.5} & \multicolumn{1}{|c}{57.6} & \multicolumn{1}{c}{44.0} & \multicolumn{1}{c}{49.9}\\
			AttentionNet & ViT & $\mathcal{L}_{SCE}$+$\mathcal{L}_{ReMSE}$ & \multicolumn{1}{|c|}{69.3} & \multicolumn{1}{|c|}{77.6} & \multicolumn{1}{|c|}{69.8} & \multicolumn{1}{|c}{62.8} & \multicolumn{1}{c}{82.1} & \multicolumn{1}{c|}{71.2} & \multicolumn{1}{|c}{70.5} & \multicolumn{1}{c}{77.7} & \multicolumn{1}{c|}{73.9} & \multicolumn{1}{|c}{57.9} & \multicolumn{1}{c}{45.5} & \multicolumn{1}{c}{51.0}\\
			\hline
		\end{tabular}
		\label{table:Ablation}
	\end{table*}
	
	\begin{figure*}[htbp]
		\centering
		\includegraphics[width=0.95\linewidth]{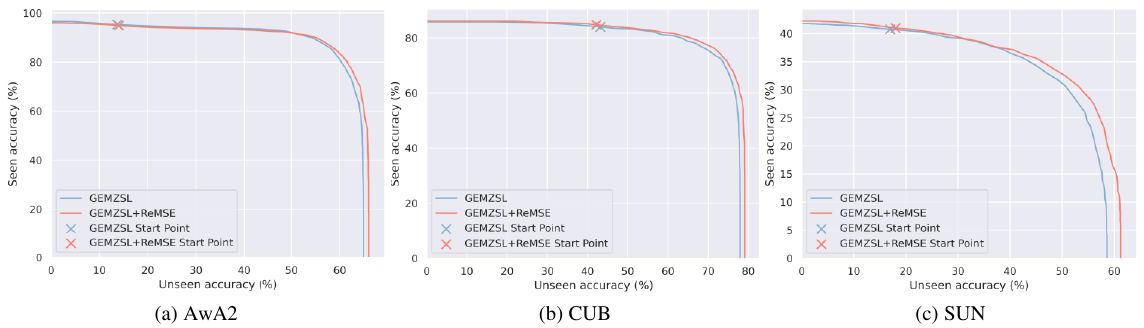}
		\caption{Visualization of Area Under Unseen-Seen Accuracy (AUSUC). Our ReMSE's AUSUC is mostly higher than the baseline model (GEMZSL).}
		\label{fig:AUSUC}
	\end{figure*}
	
	\begin{figure*}[htbp]
		\centering
		\includegraphics[width=0.95\linewidth]{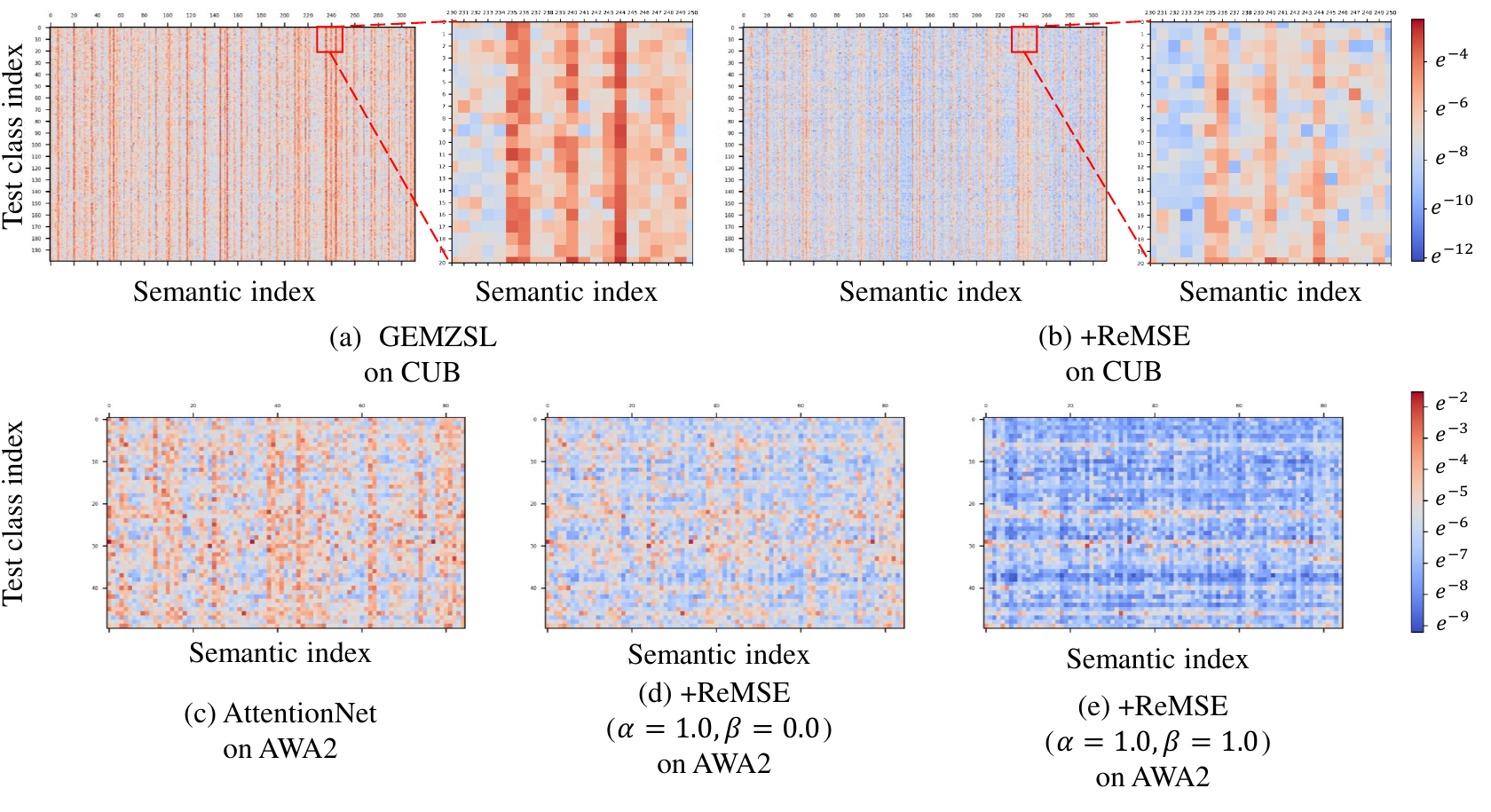}
		\caption{Class-averaged semantic error matrix on the test set of the datasets CUB and AWA2 (rescaled by logarithmic function).}
		\label{fig:OverallErrorDistribution}
	\end{figure*}
	
	To demonstrate the effectiveness of our ReMSE, we  implement various SOTA ZSL methods and evaluate our ReMSE on multiple metrics both on ZSL and GZSL settings over three popular benchmark datasets. With extensive studies, we show our ReMSE could improve various SOTA models by a significant gap as seen in Sec.~\ref{sec:SOTA}.
	We also present the imbalanced semantic regression performance in Sec.~\ref{sec:Regression}.
	Finally, we demonstrate the effectiveness of intra-class re-weighting and intra-semantic re-weighting on Sec.~\ref{sec:Ablation}.
	
	\textbf{Datasets.}
	We conduct extensive experiments to evaluate the proposed method on three ZSL benchmarks, namely (1) the coarse-grained dataset AWA2~\cite{lampert2013attribute}, one extensive animal dataset composed of 37,322 images from 50 classes (40 seen and 10 unseen) with 85-dim attributes ranged from $0$ to $100$ ($-1$ denotes missing data); (2) the fine-grained bird dataset CUB~\cite{wah2011caltech}, containing 11,788 images in 200 (150 seen and 50 unseen) classes with 312 semantics ranging from 0 to 100; (3) the fine-grained dataset SUN~\cite{patterson2012sun}, a large-scale dataset including 14,340 images from 717 classes (645 seen and 72 unseen) with 102 attributes ranging from 0 to 1.
	We divide these data into training and testing sets following~\cite{xian2018zero}, which is widely used in present methods.
	
	\textbf{Baselines \& Implementation Details.}
	We examine our ReMSE strategy on three ZSL baselines, i.e., APN, GEMZSL\footnote{It is worth mentioning that the model GEMZSL only utilizes the gaze embedding to build the attention maps, its ability to recognize unseen classes only relies on the semantics provided by the benchmark~\cite{xian2018zero}.}, and AttentionNet.
	For fair comparison, we implement APN and GEMZSL following the original training configuration, including their batch size, learning rate, sampling strategy and so on.
	For AttentionNet, we only use SCE and our proposed ReMSE loss.
	We adopt ResNet101~\cite{he2016deep} pretrained on ImageNet1K~\cite{deng2009imagenet} as the backbone. AttentionNet is optimized with a stochastic gradient descent optimizer with a learning rate of 0.0005, momentum of 0.9, and weight decay of 0.0001.
	The batch size for all datasets is set to 32.
	All the experiments were run on an NVIDIA Quadro RTX 8000 graphics card with 48GB of memory. 
	
	\textbf{Evaluation Protocols.}
	We adopt a variety of metrics for comparison.
	Specifically, for ZSL, we calculate the top-1 classification accuracy (T1) for unseen classes.
	For GZSL, we calculate three kinds of top-1 accuracies, namely the accuracy for unseen classes (denoted as $U$),  the accuracy for seen classes ($S$), and their harmonic mean:
	\begin{equation}
		H = \frac{2 \times U \times S}{U + S}.
	\end{equation}
	Besides, for GZSL, we report the performance based on  the Area Under Seen-Unseen accuracy Curve (AUSUC)~\cite{chao2016empirical}, which evaluates the degree of trade-off between $U$ and $S$ for ZSL.
	Finally, we exploit two new metrics, the mean and standard deviation of the class-averaged semantic error matrix, to evaluate the imbalanced performance on semantic regression, i.e. how well the ZSL models can fit the semantic labels and how well the error distribution is balanced, respectively.

	\subsection{Comparison with SOTAs}
	\label{sec:SOTA}
	For ZSL and GZSL tasks, we focus on embedding methods and compare our approach with classical AGEN (CVPR19), DUET (TIP19), SGAM (NeurIPS19), and more recent APN (NeurIPS20), GEMZSL (CVPR21), LSG (TIP21) and even state-of-the-art TransZero (AAAI22) and TransZero++ (TPAMI22).
	We also report the performance of various generative methods, including f-CLSWGAN (CVPR18), DCRGAN (TMM21), DisVAE~\cite{li2021generalized} (AAAI21), HSVA (NeurIPS21), CE-GZSL (CVPR21) and BSeGN~\cite{xie2022leveraging} (TNNLS22) for comprehensive reference.
	
	For imbalanced regression, we apply a Balanced MSE (CVPR22) on three embedding methods (i.e. APN, GEMZSL, and AttentionNet), which may be the first method in multi-label imbalanced regression. 
	
	The results are reported in Table~\ref{table:OverallComparison}. We highlight  three main observations: 1) ReMSE can improve the baselines consistently. For example, on the CUB dataset for ZSL task, our ReMSE endows vanilla APN, GEMZSL and AttentionNet with 0.2\%, 0.8\%, and 0.7\% performance gain, respectively, confirming that  ReMSE can effectively learn a better visual-semantic mapping. 2) On the CUB dataset, ReMSE achieves the highest score compared with AttentionNet by a considerable gap, i.e. at least 2.6\% higher than all the rest SOTAs for ZSL (Ours $80.9\%$ v.s. TransZero++ $78.3\%$), and at least $2.3\%$ (w.r.t. H) higher for GZSL (Ours $73.8\%$ v.s. GEMZSL $71.5\%$). 3) Balanced MSE may perform unstable. In some case, it may bring improvements (e.g. when it is integrated into APN for GZSL on SUN and AWA2), but in other cases, it may degrade the performance. Likewise, comparisons on the AUSUC metric (as seen in Table~\ref{table:AUSUC}) also validate that our ReMSE improves all the models with a large gap, again demonstrating the advantage of the rebalancing strategy.
	
	A visualization of the Area Under Unseen-Seen Accuracy (AUSUC) is shown in Fig.~\ref{fig:AUSUC}. We can see that our ReMSE's AUSUC is mostly higher than the baseline model (GEMZSL), which evaluates the trade-off ability of ZSL between unseen accuracy and seen accuracy. 
	
	\begin{figure*}[htbp]
		\centering
		\includegraphics[width=0.95\linewidth]{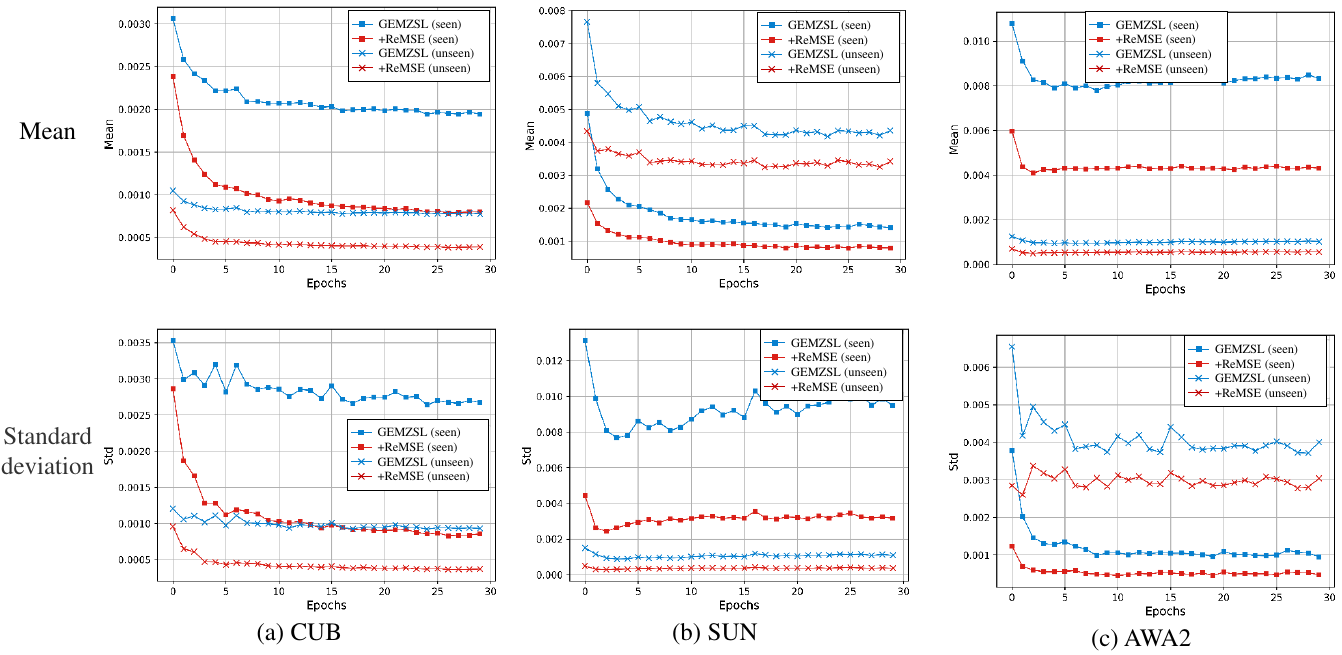}
		\caption{Mean and standard deviation of error distributions on test set. ReMSE could lead to significant drops in terms of both Mean and standard deviation of error distributions.}
		\label{fig:ErrorCurve}
	\end{figure*}

	\begin{figure*}[htbp]
		\centering
		\includegraphics[width=0.95\linewidth]{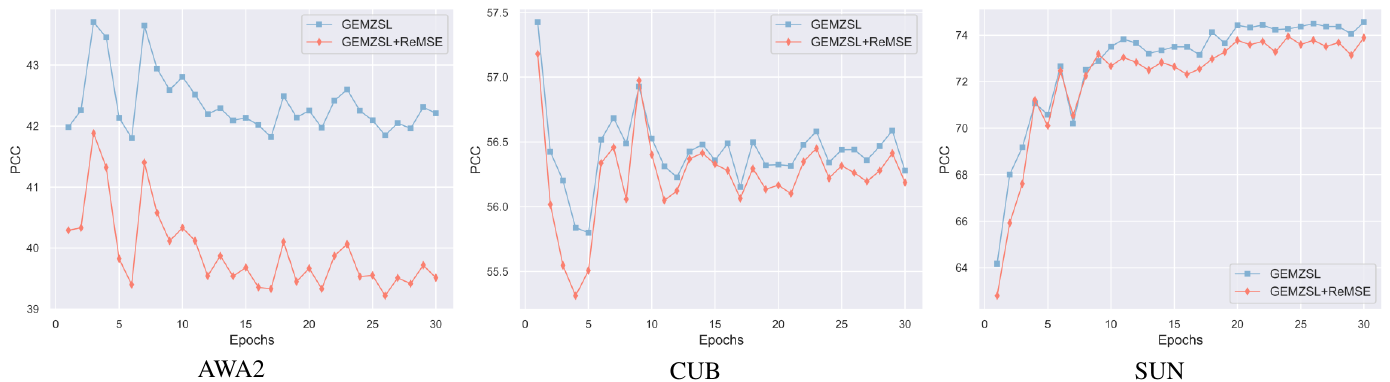}
		\caption{Visualization of the Pearson Correlation Coefficient (PCC) between semantic prediction errors and semantic label values during training. Compared with the baseline model, ReMSE makes the Pearson Correlation Coefficient drop significantly. This shows that our ReMSE method can greatly reduce the correlation between semantic prediction errors and semantic label values.}
		\label{fig:PCCCurve}
	\end{figure*}
	
	\subsection{Validation of Rebalancing Property}
	\label{sec:Regression}
	To validate that our approach can indeed rebalance errors across different classes and different semantics, we conduct several additional experiments.
	First, we visualize the variations of the error distribution on the testing set of CUB in Fig.~\ref{fig:OverallErrorDistribution}.
	Darker red means more errors, while darker blue means fewer errors. We can see that with ReMSE, the distribution of prediction errors changes from darker red to  blue overall. This clearly shows that our re-weighting could effectively suppress prediction errors without negatively affecting other well-fitting semantic regions.
	Second, conducting experiments on GEMZSL, we perform two quantitative comparisons of the mean and standard deviation of the errors distributions, as shown in Fig.~\ref{fig:ErrorCurve}.
	It is evident that once ReMSE is applied, both seen or unseen classes, the means and standard deviations drop significantly, implying that the errors are indeed balanced.
	
	We also verify that the semantic predictions of most existing models are unbalanced. Furthermore, imbalanced prediction errors are often associated with semantic labels. To demonstrate that our ReMSE can reduce undesired correlations, we visualize the Pearson Correlation coefficient (PCC) between semantic prediction errors and semantic label values, as shown in Fig.~\ref{fig:PCCCurve}. We can observe that our ReMSE leads that the PCC drops significantly compared to the baseline model, which indicates that our ReMSE can indeed greatly reduce the linear relationship between semantic prediction error and semantic label value, thereby alleviating the  imbalanced semantic prediction issue.

	\begin{figure*}[htbp]
		\centering
		\includegraphics[width=0.95\linewidth]{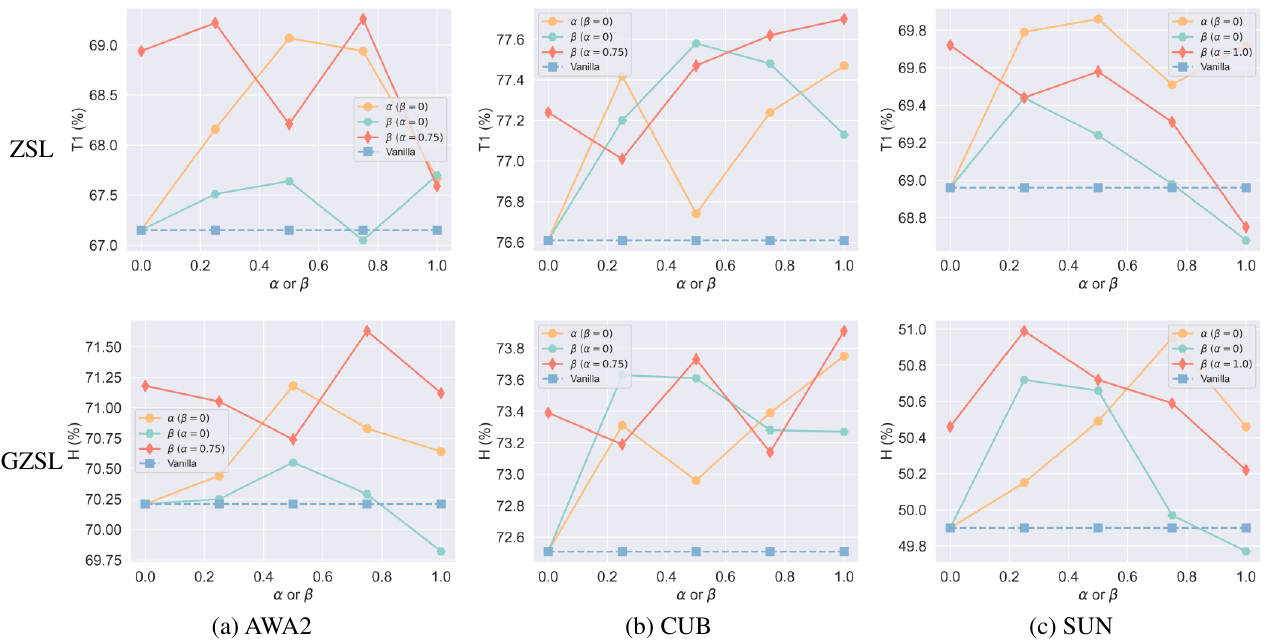}
		\caption{
			Effects of re-weighting hyper-parameters $\alpha$ (class-level) and $\beta$ (semantic-level). The vanilla model is AttentionNet with ViT backbone.
		}
		\label{fig:ablationUpdated}
	\end{figure*}

	\subsection{Ablation Study}
	\label{sec:Ablation}
	\subsubsection{Component analysis}
	\label{sec:ComAna}
	We conduct ablation studies to verify the effectiveness of our approach.
	Table~\ref{table:Ablation} shows the impact of each component.
	We first use SCE loss to train a model that only contains global branch (GB) or attention branch (AB).
	Next, we fuse these two branches as the full AttentionNet.
	After that, our ReMSE are added to our AttentionNet.
	From the table, we could get three conclusions: (1) The attention branch might cause some degradation. For example, the model without AB could perform better than the model without GB on AWA2 and SUN. (2) The first three column indicates combining global and attentive features could improve the expressiveness of features, and allow models predict semantics more accurate.
	(3) Remarkly, Our proposed ReMSE improves the T1 of ZSL over the model trained by SCE by 1.6\% (AWA2), 0.7\% (CUB) and 0.4\% (SUN), respectively, and the harmonic mean accuracy (H) of GZSL by 0.3\% (AWA2), 0.9\% (CUB) and 1.5\% (SUN), respectively.
	This influence verifies the effectiveness of our ReMSE that does not only decrease the mean of errors but also reduce the variance of errors.
	
	Besides, we also evaluate the effect of the recently popular Vision Transformer (ViT)~\cite{han2022survey}.
	We can observe that our ReMSE consistently improves both the ZSL and GZSL models.
	Moreover, the ViT-based variants are better than ResNet-based variants in SUN that means the features extracted by ViT are extremely good at scene classification.

	\subsubsection{Sensitivity Analysis}
	\label{sec:Sensitivity}
	We take the three datasets to analyze the Sensitivity of the hyperparameters $\alpha$ and $\beta$ used in the rebalance method.
	As shown in Fig.~\ref{fig:ablationUpdated}, a proper $\alpha$ or $\beta$ can bring improvement of T1, which appears to be higher than the vanilla method. 	In addition, re-balancing at both the class-level and semantic-level are more effective than only re-balancing a single head in almost all the cases.

	\begin{figure*}[htbp]
		\centering
		\includegraphics[width=0.9\linewidth]{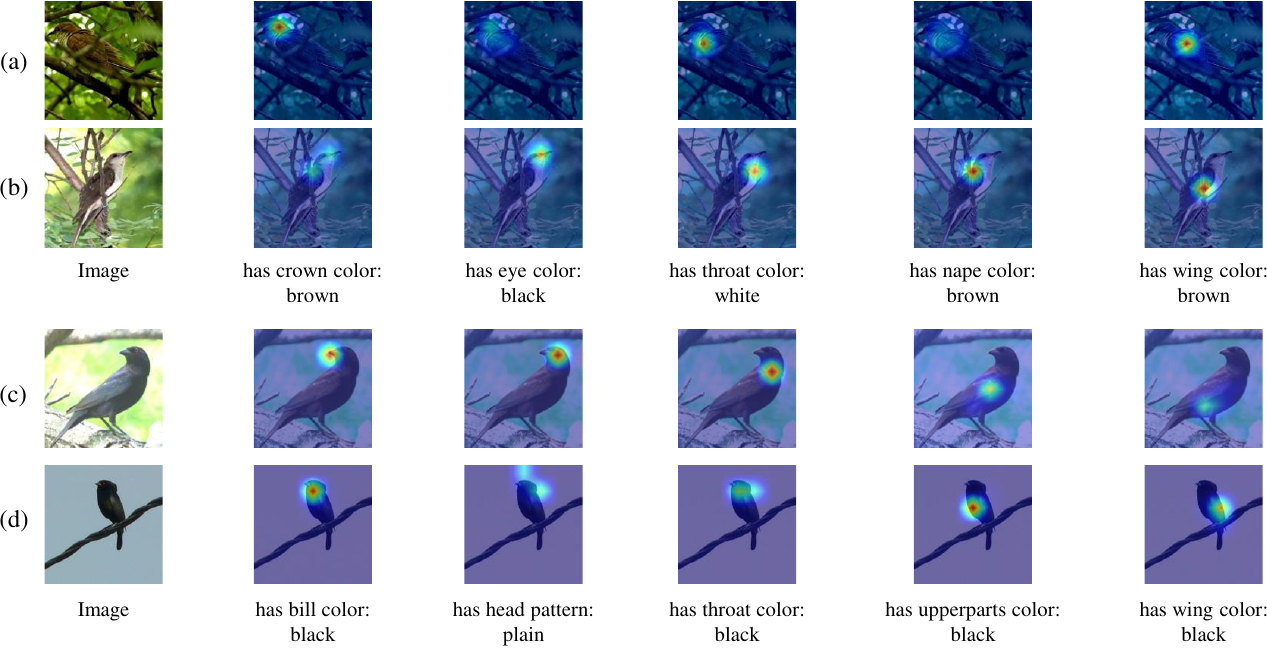}
		\caption{Visualization of attention maps produced by our AttentionNet according for different semantics of unseen images on the dataset CUB. The attention map has a resolution of $7 \times 7$, and is reshaped into $224 \times 224$ to match the image size.}
		\label{fig:attention_vis1}
	\end{figure*}
	
	\begin{figure*}[htbp]
		\centering
		\includegraphics[width=0.95\linewidth]{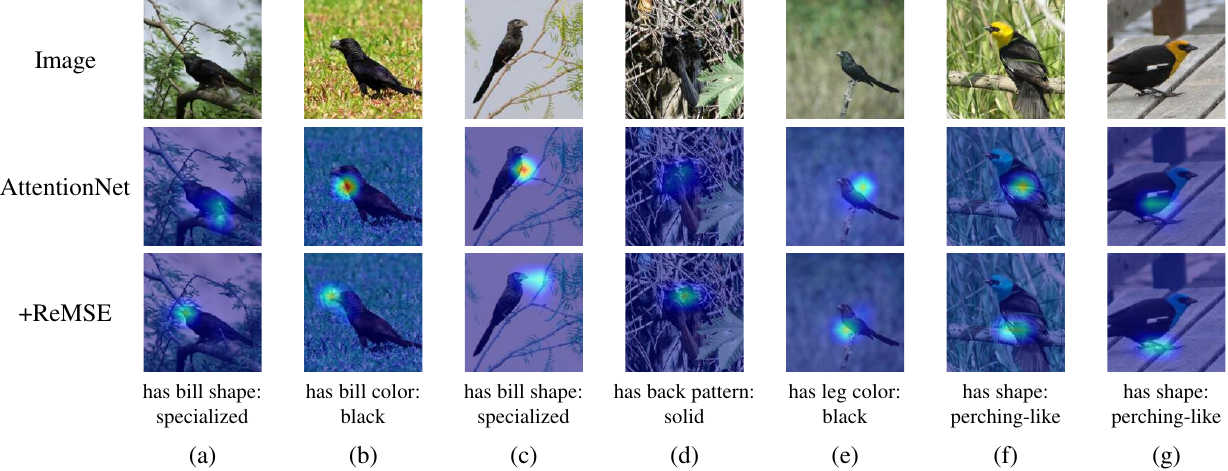}
		\caption{
			The effect of our ReMSE algorithm on the attention map. The settings are the same as in Fig.~\ref{fig:attention_vis1}, but the model results in a more accurate attention map for difficult semantics.}
		\label{fig:attention_vis2}
	\end{figure*}
	
	\subsection{Visualization of Attention}
	We also visualize the attention map of our AttentionNet to qualitatively verify its effectiveness as shown in Fig.~\ref{fig:attention_vis1}.
	The figure shows the results of different attention maps according to various semantics.
	Obviously, our AttentionNet adaptively detects the semantic regions that are beneficial for prediction.
	For example, when the semantics are related to crown, eye and bill, the attention is distributed to heads of birds.
	When the semantics are related to wing or upper-parts, the attentive regions become the bodies.

	Moreover, we also verify the effectiveness of our ReMSE for the attention, as shown in Fig.~\ref{fig:attention_vis2}.
	We can see that with the help of our ReMSE the model corrects its out-of-focus regions.
	For instance, for the semantics related to bills, the model incorrectly focuses on (a) the legs, (b) the throat, and (c) the chest.
	But with the help of ReMSE, it exactly focuses on  bills.
	Generally, these figures illustrate that our ReMSE plays a key role in   predicting accurately hard semantics.

	\section{Conclusion}
	In this work, we address the zero-shot learning problem from a brand new perspective of imbalanced learning. We propose the ReMSE strategy and focus on re-balancing the imbalanced error distribution across different classes and different semantics. We set out a series of analyses both theoretically and empirically to validate the rationale of ReMSE in ZSL. Extensive experiments on three benchmark datasets show that ReMSE can consistently improve the three baselines, achieving competitive performance even compared to those sophisticated ZSL methods.

	\section*{Acknowledgments}
	This work was partially supported by ``Qing Lan Project" in Jiangsu universities, National Natural Science Foundation of China under No. 62106081, Research Development Fund with No. RDF-22-01-020, and Jiangsu Science and Technology Programme  under No. BE2020006-4.

	
	\newpage
	
	\bibliography{ReZSL}
	\bibliographystyle{IEEEtran}

	\section{Appendix}
	\subsection{Normalization of MSE}
	\begin{shaded}
		\begin{proposition}
			For any regressor with the MSE loss and parameter $\mathbf{W}$, given any example $(\mathbf{x}_i,\mathbf{s}_i,y_i)$, when the model tends to be optimal, we have $\|\tilde{\mathbf{s}}_i\|_{2} \rightarrow \|\mathbf{s}_i\|_2 \cos \theta$, where $\theta$ is the angle between the semantic value and its predicted value. 
		\end{proposition}
	\end{shaded}
	\begin{proof}
		Recalling the formulation in Eqn.~(\ref{eq:MSE}), the MSE loss can be expanded  as follows:
		\begin{align}
			\mathcal{L}_{MSE} = \frac{1}{N} \sum_{i = 1}^N (\|\tilde{\mathbf{s}}_i\|_2^2 - 2\|\tilde{\mathbf{s}}_i\|_2 \|\mathbf{s}_i\|_2 \cos \theta + \|\mathbf{s}_i\|_2^2). \nonumber
		\end{align}
		By directly calculating the gradient of the MSE loss with respect to the parameter $\mathbf{W}$, we obtain
		\begin{align}
			\label{eq:MSEgrad}
			\frac{ \partial \mathcal{L}_{MSE}  }{ \partial \mathbf{W} } & = \frac{1}{N} \sum_{i = 1}^N \frac{ \partial \mathcal{L}_{MSE}} { \partial \|\tilde{\mathbf{s}}_i \|_2 } \frac{ \partial \|\tilde{\mathbf{s}}_i \|_2}{\partial \mathbf{W} } \nonumber \\
			& = \frac{1}{N} \sum_{i = 1}^N 2(\|\tilde{\mathbf{s}}_i\|_2 -  \|\mathbf{s}_i\|_2 \cos \theta) \frac{ \partial \|\tilde{\mathbf{s}}_i \|_2 }{\partial \mathbf{W} }. \nonumber
		\end{align}
		Hence, when the gradient tends to be $0$, the magnitude of $\frac{ \partial \|\tilde{\mathbf{s}}_i \|_2 }{\partial \mathbf{W} }$  becomes very small, then we have $\|\tilde{\mathbf{s}}_i\|_2 \rightarrow \|\mathbf{s}_i\|_2 \cos \theta$.
	\end{proof}
	
	\begin{remark}
		This proposition shows that the original MSE may not precisely measure how well the semantics fit.
		We can see that even if $\|\tilde{\mathbf{s}}_i\|_2 \rightarrow \|\mathbf{s}_i\|_2 \cos \theta$,  $\tilde{\mathbf{s}}_i$ and $\mathbf{s}_i$ may still generate a big angle. Therefore, the predicted and ground-truth semantics can still be very inconsistent. 
		Thus, we normalize both semantic labels and predictions in MSE to obtain the Normalized MSE $\mathcal{L}_{NMSE}(\tilde{\mathbf{s}}_i,\mathbf{s}_i) = \mathcal{L}_{MSE}(\tilde{\mathbf{t}}_i,\mathbf{t}_i)$, which proves fairly compatible with $\mathcal{L}_{SCE}$, since the $\mathcal{L}_{NMSE}= 2- \cos\theta(\tilde{\mathbf{s}}_i, \mathbf{s}_i)$.
	\end{remark}
	
	\subsection{Proof of Convergence of ReMSE}
	\begin{shaded}
		\begin{theorem}
			\label{theorem:rmes} Given two classes of data $\{(x,c_1),(y,c_2)\}$, which are located in a 1-dimensional space, and  $\tilde{x}$ and $\tilde{y}$ are the prediction values of the semantic values $x$ and $y$, respectively. The loss function become
			\begin{equation}
				\mathcal{L}_{ReMSE} = w(x - \tilde{x})^2 + v(y - \tilde{y})^2,
			\end{equation}
			where $w = p_{1 1}q_{1 1}$ and $v = p_{2 1}q_{2 1}$.
			Minimizing $\mathcal{L}_{ReMSE}$ by gradient descent will minimize prediction errors for each class as well as the variance of prediction errors.
		\end{theorem}
	\end{shaded}

	\begin{proof}
		For the $t$-th iteration, we have $w_t \ge 1,v_t = 1$ or $w_t = 1, v_t \ge 1$. We know that the prediction errors for the two samples in the $t$-th iteration are $(\tilde{x}_t - x)^2$ and $(\tilde{y}_t - y)^2$. 
		Without loss of generality, let us assume that $(\tilde{x}_t - x)^2 < (\tilde{y}_t - y)^2$, then we have $w_t = 1$ and $v_t > 1$. We update $\tilde{x}$ and $\tilde{y}$ by gradient descent with a stepsize of $r$ (a small positive constant), and get
		\begin{align}
			\tilde{x}_{t + 1} &= \tilde{x}_t - 2 r w_t (\tilde{x}_t - x), \\
			\tilde{y}_{t + 1} &= \tilde{y}_t - 2 r v_t (\tilde{y}_t - y).
		\end{align}
		Therefore, we get a new prediction error as follows
		\begin{align}
			(\tilde{x}_{t + 1} - x)^2 &= (1 - 2 r w_t)^2 (\tilde{x}_t - x)^2, \\
			(\tilde{y}_{t + 1} - y)^2 &= (1 - 2 r v_t)^2 (\tilde{y}_t - y)^2.
		\end{align}
		At this point, since $(1 - 2 r w_t)^2 > (1 - 2 r v_t)^2$, we have
		\begin{align}
			& (\tilde{y}_{t + 1} - y)^2 - (\tilde{x}_{t + 1} - x)^2 \nonumber \\
			& = (1 - 2 r v_t)^2 (\tilde{y}_t - y)^2 - (1 - 2 r w_t)^2 (\tilde{x}_t - x)^2 \nonumber \\
			& < (1 - 2 r w_t)^2 \left((\tilde{y}_t - y)^2 - (\tilde{x}_t - x)^2 \right).
		\end{align}
		Since $(1 - 2 r w_t)^2 < 1$,  after a gradient descent, the difference in prediction error between the two classes becomes smaller, that is,  these losses become more balanced. The weight $w_t$ and $v_t$ are continuously adjusted until $(\tilde{x}_t - x)^2 = (\tilde{y}_t - y)^2$, at which point we have $w_t = 1$,$v_t = 1$ and $(\tilde{x}_{t + 1} - x)^2 = (\tilde{y}_{t + 1} - y)^2$.
		
		Finally, it is worth noting that after the $t$-th iteration, we have $(\tilde{x}_{t + 1} - x)^2 \le (\tilde{x}_t - x)^2$ and $(\tilde{y}_{t + 1} - y)^2 \le (\tilde{y}_t - y)^2$. Namely, the prediction error for each class decreases as the iteration progresses. Our ReMSE algorithm will adjust the weights so that they end up being balanced. Ideally, when our loss is minimized, the average prediction error for each class is roughly equal.
	\end{proof}
	
	It is worth noting that, considering the separability of the loss function along the class dimension and the semantic dimension, if the optimization task of the loss function $\mathcal{L}_{ReMSE}$ is regarded as a $|\mathcal{C}^u| \times d_{s}$ independent 1-dimensional regression problem, then the above conclusion can be easily extended to multi-class and multi-semantic situations.
	
	\subsection{Balanced MSE}
	The Balanced MSE~\cite{ren2021balanced} also aims to handle imbalanced error distributions by assuming that the distribution of sample is balanced. Its batch-based formulation is
	\begin{equation}
		\label{eq:BalancedMSE}
		\mathcal{L}_{balMSE} = -\log \frac{e^{-\|\tilde{\mathbf{t}}_{i}-\mathbf{t}_{i}\|^2_2/\sigma } }{\sum_{\mathbf{t}_{j} \in B_{\mathbf{t}}} e^{-\|\tilde{\mathbf{t}}_{i}-\mathbf{t}_{j}\|^2_2/\sigma} },
	\end{equation}
	where $B_{\mathbf{t}}$ is a batch of normalized semantic labels and $\sigma$ is a learnable parameter. It can be viewed as $\mathcal{L}_{NMSE}$ with a regularization to balance sample distribution.
	However, as our find that the imbalanced error distribution in ZSL is caused by the imbalanced semantic values, rather than the sample size, the Balanced MSE does not perform well in the ZSL setting.

\end{document}